%% file: nle.tex
\documentclass[11pt]{nle}

\usepackage[pdftex]{graphicx}
\usepackage[pdftex]{color}

\title[Extraction of Templates using SeqBDD]{Extraction of Templates from Phrases \\
Using Sequence Binary Decision Diagrams}

\author[Hirano, Tanaka-Ishii and Finch]{
  \hspace*{-1cm}D.\ns H\ls I\ls R\ls A\ls N\ls O\\
         Graduate School and Faculty of Information Science and Electrical Engineering, Kyushu University, Japan. \\
         E-mail: {\tt hirano@limu.ait.kyushu-u.ac.jp}
       \and 
       K.\ns T\ls A\ls N\ls A\ls K\ls A\ls -\ls I\ls S\ls H\ls I\ls I \\
       Research Center for Advanced Science and Technology, University of Tokyo, Japan.\\
       E-mail: {\tt kumiko@cl.rcast.u-tokyo.ac.jp} 
       \and 
       A.\ns F\ls I\ls N\ls C\ls H\\
National Institute of Information and Communications Technology  Kyoto, Japan. \\
       E-mail: {\tt andrew.finch@gmail.com}}

\received{13 March 2017; revised August 2017; March and May 2018}

\pagerange{\pageref{763}--\pageref{795}}
\pubyear{2018}

\usepackage{graphicx}
\usepackage{amssymb}
\usepackage{amsmath}
\usepackage{color}
\usepackage{wrapfig}
\usepackage{url}
\usepackage{ascmac}

\usepackage{color}
\usepackage{algorithm}
\usepackage{algorithmic}
\usepackage{here}
\let\bibhang\relax
\usepackage[round]{natbib}
\usepackage{eclbkbox}
\usepackage{setspace}

\newcommand{\secref}[1]{\S\ref{#1}}
\newcommand{\figref}[1]{Fig.\ref{#1}}
\newcommand{\tabref}[1]{Table \ref{#1}}
\newcommand{\algref}[1]{Algorithm \ref{#1}}
\newcommand{\apdref}[1]{Appendix \ref{#1}}
\begin{document}

\label{firstpage}
\maketitle

\begin{abstract}
The extraction of templates such as ``regard X as Y'' from a set of
related phrases requires the identification of their internal
structures.  This paper presents an unsupervised approach for extracting
templates on-the-fly from only tagged text by using a novel relaxed
variant of the Sequence Binary Decision Diagram (SeqBDD). A SeqBDD can
compress a set of sequences into a graphical structure equivalent to a
minimal DFA, but more compact and better suited to the task of template
extraction.  The main contribution of this paper is a relaxed form of
the SeqBDD construction algorithm that enables it to form general
representations from a small amount of data. The process of compression
of shared structures in the text during Relaxed SeqBDD construction,
naturally induces the templates we wish to extract. Experiments show
that the method is capable of high-quality extraction on tasks based on
verb+preposition templates from corpora and phrasal templates from short messages
from social media.
\end{abstract}

\section{Introduction}

The extraction of frequently appearing templates such as ``regard X as
Y'' or ``Magnitude X earthquake at Y'' is a challenging problem since
it requires an analyzer to capture the structure of the frequently
appearing phrases, and detect which part should be represented by the
slots. In this article, a template is defined as a frequently
  co-occurring subsequence of words containing one or more slots, where a
  slot is replaceable by another word sequence.  In other words, a
  template in this article is a sequence of elements of length longer
  than one, where an element is either a word or a slot: at least
  one element is a slot, the other elements are words. 
  The template itself, or the
  word sequences filling the slots in this article are not necessarily 
  limited to linguistic NP, VPs and PPs.

Classically, such templates have been considered in relation to
collocations such as in
\citep*{Smadja:1993:RCT:972450.972458}. Collocations without slots are
readily extractable by finding $n$-grams that form atomic
chunks. Therefore researchers have been turning their attention to the
more difficult and general task of extracting templates with
slots. The problem has been addressed in part in the form of the
extraction of verb structures.  \citet{baldwin-handbook10} showed the
possibility of MWE (multi-word expressions) extraction by only using $n$-grams and co-occurrences, without using
a parser.

Since templates are related to the grammatical structure of the phrases,
one possibility is to use a parser. However, the kinds of text and
languages that can be processed by a parser are strictly limited, since
they are generally built from clean annotated corpora. In addition parse
trees can be deep, and the recursive nature of parse trees means there are
no limits on the depth of the structures involved, potentially leading
to a heavy computational burden.


Templates, as defined at the beginning of the article,
are commonly found in the social media: social media text
is usually short (around 140 characters in the case of
Twitter) and many distributers use automatic bots to communicate the
recent database updates.  We consider the extraction of these
templates as one target application of our approach.  Social media
text can cause issues even with a well-trained parser, since their
word sequences are often non-grammatical.  Furthermore, the
grammatical structures found in this type of text are typically
shallow making a full-blown parser less appropriate for this task than
a technique that focuses on the shallow structure directly.  It may be
possible to develop an unsupervised parsing technique to handle social
media text, but a general high quality method for unsupervised parsing
is still being sought~ \citep*{Headden:2009:IUD:1620754.1620769}.  One
possible explanation for this is the difficulty inherent in analyzing
the structure recursively. Another related field, shallow
parsing (\citep*{Abney91parsingby,DBLP:journals/corr/HuangXY15}) is
well-established and devoted to obtaining shallow linguistic
analyses. These techniques typically segment a sentence into
grammatical structures. While pursuing this line may have led to
fruitful results, instead, we chose to try an alternative
unsupervised method, a graphical method that produces a flat
analytical structure, as opposed to the potentially unlimited depth of
the structures that can be created by a full parser.

%

Our proposed unsupervised method extracts 
templates through the construction of a structure equivalent to a
minimum deterministic finite state automaton (minDFA)  \citep*{Daciuk00}.
Our implementation uses Sequence Binary Decision Diagrams (SeqBDDs) to
allow the automata to be more memory efficient than traditional minDFA
(around 20\% was reported in  \citep*{denzumi16}), and furthermore permits
the graph to be constructed on-the-fly. 

The technique we propose facilitates the mining of text from the
social media, but can also find application in the more general field
of lexicography and corpus linguistics.  Patterns of word usage can
play an important role for enriching dictionaries, and moreover serve
language learners as in the lexical templates found in the Collins
Cobuild English Dictionary~ \citep*{COBUILD-nouns}. As a means of determining
such templates, the key word in context (KWIC) technique is often
employed to rapidly display words in their textual context, although
the template identification must be done laboriously by hand.

The primary contribution of this paper is to propose an algorithm for
automatically constructing graphical representations of text designed
specifically for automatic template extraction. The SeqBDD (or
minDFA) is the starting point of this work, and we show empirically
that it is capable of generating templates with high precision, but
suffers from low recall. We propose a relaxed algorithm for SeqBDD
construction to overcome this fundamental shortcoming. Usually,
attempting to raise recall leads to a lowering of precision, but in this
work, the relaxation process increases the number of templates being
extracted while at the improving their quality, leading to gains 
in both precision and recall.

\section{Related Work}

We are interested in
extracting the templates as defined at the beginning of Introduction.
Although word sequences filling the slots in this article are not
necessarily limited to linguistic units such as NP, VPs and PPs, we
want the results to be meaningful linguistically. The templates
studied in this article can be considered multi-word expressions
(MWEs) with slots, where MWEs are coined \citep*{baldwin-handbook10},
as ``the split configuration of the verb-particle constructions
(VPCs)''.  A detailed summary of the acquisition of multi-word expressions is
well set out in this reference. In brief, active research in MWEs was
triggered by the seminal paper of \citet*{sag2002multiword}. In the
work that followed, various attempts have been made to acquire MWEs,
such as co-detection of English and Japanese templates by statistical
methods \citep*{Kita94}, or the extraction of verbs and noun-type
idiom candidates \citep*{Fazly06automaticallyconstructing}.
\citet{Duan:2006:BAM:1273073.1273096} reapply a method to extract
frequent patterns from DNA sequences.  Since the basic technique is
based on finding longest common subsequences, the method does not
handle gaps.  Several different methods for extracting MWEs using a
bilingual corpus have also been proposed
\citep*{caseli:2010:lre,ZarrieB:2009:ETC:1698239.1698245}.

In more recent work, \citet{nagyt-vincze:2014:MWE} focus on the
extraction of verb-particle combinations, each of which has a verb and
a particle. 
Their method can handle gaps since it is a syntax-based method that uses a parser,
but this requires a fully annotated gold standard.
A
more recent work attempts to extract MWEs without wildcards.
\citet{Tu:2012:SOM:2387636.2387648} study the method to detect 
phrasal verbs using machine learning techniques.  Although slots are considered
 in this article, the application is limited to 6
verbs and 19 prepositions.  Some recent studies attempt to extract
from grammatically structured data, such as treebank and parsed data.
\citet{sangati2015multiword} extract multiword expressions from
frequently appearing fragments in the treebank.  Since the method is able to exploit a
treebank, the accuracy of the method is higher than the unsupervised
methods. Another study \citep*{martens:2010:POSTERS} attempts to find
frequent subtrees from parsed data.  Our approach differs from these in that it
assumes no grammatical structure in the input.

From the opposite perspective, the proposed method is aimed at
discovering the underlying structure in related phrases in an unsupervised
manner, and as such it is broadly related to work in the field of
unsupervised grammar induction.

In the field of natural language processing, the induction of grammars
more complex than regular grammars has been actively studied.  In
 \citep*{Klein:2004:CIS:1218955.1219016}, an unsupervised dependency
parser DMV was proposed which successfully induced structure from a
modest amount of data. The success of this early approach stimulated
further research in the field.
 \citet{Headden:2009:IUD:1620754.1620769} introduced valence frames
and lexical information into their induction process, giving rise to a
substantial improvement in performance
 \citep*{gimpel-smith:2012:NAACL-HLT2}. The method of
 \citet*{Klein:2004:CIS:1218955.1219016} was improved in
 \citep*{gimpel-smith:2012:NAACL-HLT2} by introducing concave models
that could be optimized without the problem of local minima.  However,
at present the accuracy of the state-of-the-art unsupervised methods
evaluated on Penn Treebank text
 \citep*{Marcus:1993:BLA:972470.972475} is below 70\% when the relative to the
state-of-the-art supervised methods achieve over 93\%
 \citep*{gimpel-smith:2012:NAACL-HLT2,Klein:2004:CIS:1218955.1219016}.
Therefore,
it may be difficult to replace current supervised parsers with
unsupervised parsers to use for template extraction.

A template considered in this work
has the basic structure of a regular expression (that we
will model using automaton-related data-structures), and therefore
next we summarize previous work related to regular expression
induction that directly concerns template induction.  The work can be
roughly partitioned into three: heuristics, regular grammar induction,
and methods related to finite state automata.  Many have performed
template induction based on heuristics, such as
 \citep*{PrefixSpan,cui04,kim2010,Han:2000:MFP:342009.335372}, just to
mention a few.  Most of the approaches are original but none are
theoretically grounded.  The remainder of the previous work related to
regular expressions presented here has some mathematical
framework as a basis.

Regular grammar induction algorithms are able to extract a regular
grammar from a set of phrases with similar structure. 
These techniques have had wide application to sequences in general
including the extraction of DNA patterns.  \citet{fernau09}
suggested a simple bottom-up regular expression induction strategy that
we replicated and applied to natural language. Unfortunately it
resulted in a huge number of trivial regular expressions. Moreover,
it is not straightforward to apply the method to template extraction since
it has no mechanism for slot creation.
Most of the previous work discussed so far is based on prefix sharing
of similar phrase instances. This however generates huge amount of
template candidates that have similar infix and suffix structures.  The
point of phrasal template induction is to identify shared common
structures anywhere in the phrases, and abstract them into slots.


Template extraction has a broad range of potential applications.  One
potential application is to support lexicographers in
building/enriching dictionaries for humans and machines.  The
application can take a dynamic form, for example, by extending the
keyword in context (KWIC) system \citep*{pattern-grammar}, which shows
the words immediately preceding and following the node word, up to a
predetermined length.  In fact, our work was partly motivated by an
idea presented in ``pattern grammars'' \citep*{pattern-grammar}.  The
authors assembled the Collins Cobuild Grammar Templates dictionary
\citep*{COBUILD-verbs,COBUILD-nouns}, and claim that texts are formed
of {\em patterns} defined as word sequences that may contain slots
(and we use the term {\em template} to represent this concept hereafter). The
patterns allow for parsing any sentence using pattern grammar, as each
word in the sentence will occur with its own
pattern. \citet{pattern-grammar} shows this under the heading of
`pattern flow'.  Although the Collins Cobuild English Dictionary was
constructed manually, they suggest the possibility of automating the
procedure with an extended form of the KWIC system.  This work takes a first
step to answer this challenge, by means of the SeqBDD.  This work was
motivated by this fundamental work, but the templates we extract are
different from patterns of this Pattern Grammar, in which patterns are
defined as sequences of phrase or clause types (e.g. 'that-clause',
'noun phrase') that are dependent on specific verbs, nouns or
adjectives.

Another genre of application, as mentioned in the previous section,
relates to social media, which has abundant repetitive patterns and
shallow structure.  \citet{Baldwin13hownoisy} shows these common
characteristics threading through social media, through building and analyzing
a corpus consisting of various types of social media. This suggests how
patterns could be one possible key to information extraction.  The
identification of patterns would also create a means of compressing the data. 
This paper will
later study the proposed method in the context of these two potential
applications: the verb+preposition template extraction as an example
towards an extended KWIC method, and pattern acquisition from social
media.

\section{The Sequence BDD}

Given a set of linguistic phrases with a similar structure, we want to extract
the common structure as a {\em template}: a sequence of words and slots.
The first attempt of doing this made use of a Trie structure.
The Trie however, does not allow for sharing the infix and suffix.
Broadly, for template extraction, we are interested in representations
that also allow for the sharing of common structures within infixes
and suffixes.

Among the different data structures that allow this type of sharing,
we chose the SeqBDD (Sequence Binary Decision Diagram) as our basis.
The SeqBDD is a descendant of two different formal methods: of Binary
Decision Diagrams (BDD) and acyclic Deterministic Finite state
Automata (DFA)\footnote{ The reason why DFA are acyclic is that all
  string sets we need to model are finite. Similarly, the SeqBDD is also
  acyclic.}.

Therefore, the explanation of SeqBDD has two aspects: SeqBDD as
one descendant of BDD; and the other in contrast to DFA.  We provide the
former in \secref{sec:seqbdd_vs_bdd}, and the latter in \secref{sec:vs}.  The basis of the SeqBDD can be
found in \citep*{Loekito:2010:BDD,denzumi16}, and the definitions
and algorithms necessary to replicate our method are all provided in
this article.

\begin{figure}[t]
\centering
\includegraphics[width=0.7\columnwidth]{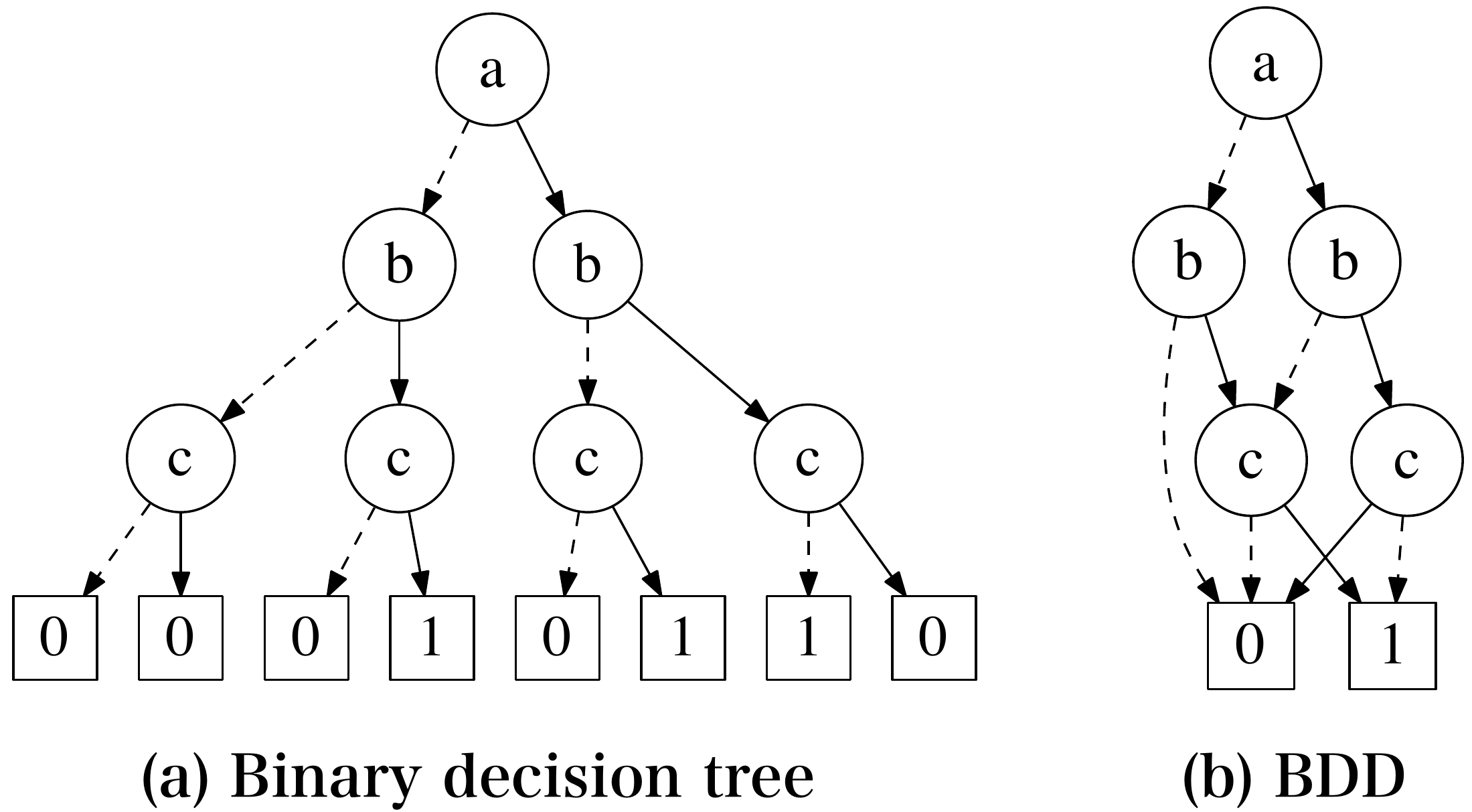}
\caption{A binary decision tree and corresponding BDD structures representing the
Boolean function $ab\bar{c} \vee a\bar{b}c \vee \bar{a}bc$.}
\label{fig:bdtree_samples}
\end{figure}

\subsection{SeqBDD as a descendant of BDD}
\label{sec:seqbdd_vs_bdd}

SeqBDDs are descendants of Binary Decision Diagrams (BDDs)
\citep*{Bryant:1986:GAB:6432.6433} that are graphical structures able
to express Boolean functions.  For example, consider the Boolean
function `$ab\bar{c} \vee a\bar{b}c \vee \bar{a}bc$'. The function can
be represented as the binary decision tree structure shown in
\figref{fig:bdtree_samples}(a).  Solid edges in the graph represent
truth (1-edge), and dashed edges represent falsehood (0-edge).  The
leaf nodes represent final truth and falsehood as 1 and 0,
respectively.  It is clear from the figure that the 
tree contains many common substructures.  For example, a node labeled
`$c$' is connected to the 0- and 1-terminal nodes multiple times.  The
study of BDD has tackled the elimination of such redundancy in binary
decision trees \citep*{Bryant:1986:GAB:6432.6433,knuth09}.

The minimization of the representation is conducted by exhaustively
applying two reduction operations for merging identical
subgraphs: 1)
node sharing (share all equivalent subgraphs), and 2) node deletion
(delete all redundant nodes where both outgoing
edges lead to equivalent subgraphs)
 \citep*{Bryant:1986:GAB:6432.6433,knuth09}.  
Equivalent subgraphs are defined recursively: two nodes are in
equivalent subgraphs if their labels are identical and the heads of
their outgoing edges are also in the equivalent subgraphs.  For
example, in \figref{fig:bdtree_samples}(a) the central two nodes
labeled `$c$' are both in equivalent subgraphs that extend down to
the leaf nodes. These subgraphs are merged into the left `$c$' node at
the bottom in \figref{fig:bdtree_samples}(b) using the node sharing
operation.

As for the node deletion, the left-most node labeled `$c$' in
\figref{fig:bdtree_samples}(a) is deleted using the node deletion
operation 2) since both outgoing edges lead to the 0-terminal node.  Its
incoming edge (from node `$b$') is redirected to the shared equivalent
subgraph of its children (the 0-terminal node).  It is clear that the
graph in (b) is considerably more compact than (a).

\begin{figure}
\centering
\includegraphics[width=0.75\columnwidth]{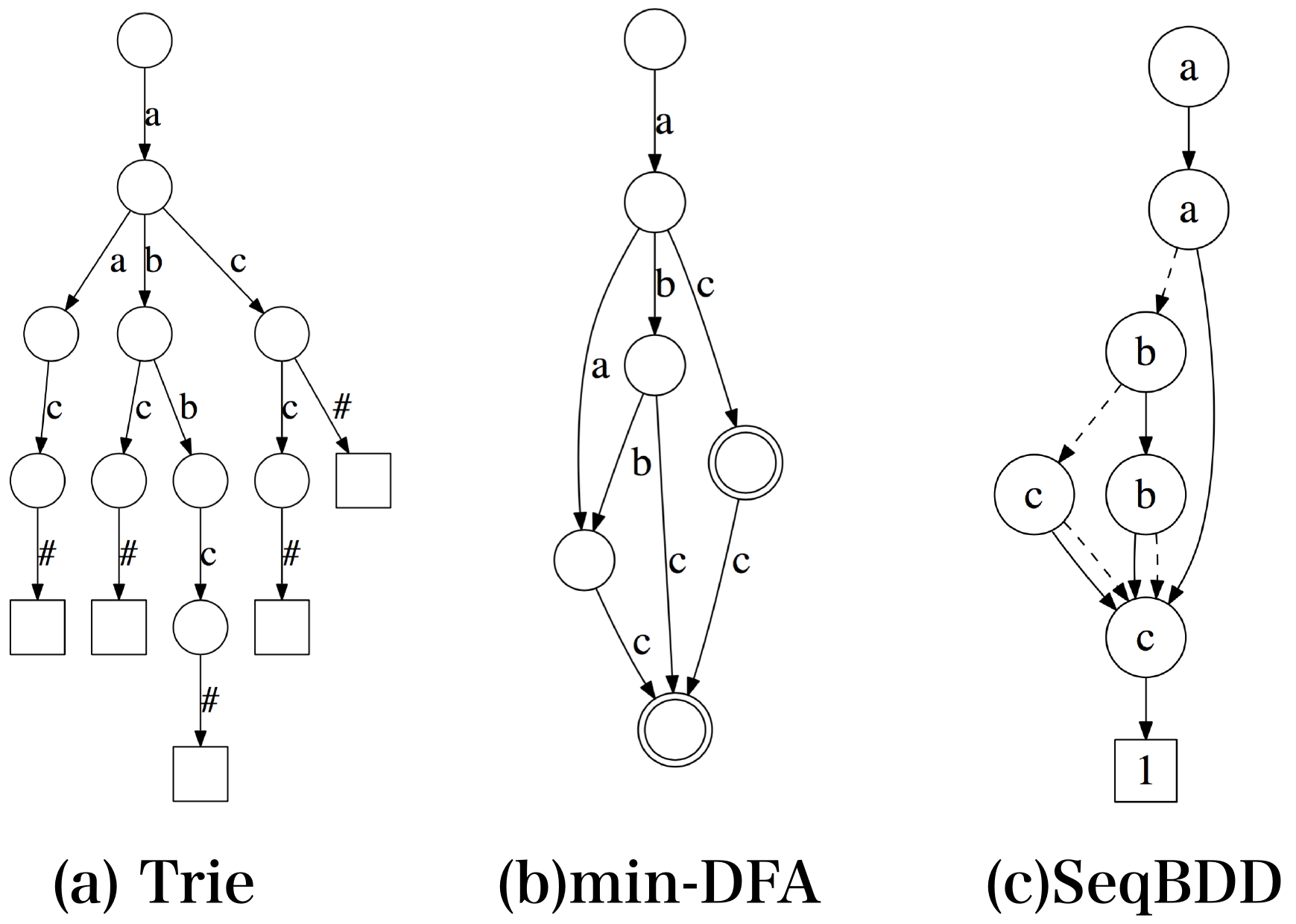}
\caption{Equivalent Trie, minDFA and SeqBDD representing: $\left\lbrace ac, abc, aac, acc, abbc \right\rbrace $. }
\label{fig:try_seqbdd}
\end{figure}


A SeqBDD \citep*{Loekito:2010:BDD} is a BDD dedicated to representing
a set of sequences. A set of sequences is treated as a Boolean
function; the function being True if the sequence is a member of the
set, and False otherwise. Since the number of False members are
infinite, they are not explicitly shown in SeqBDD.  Paths along the
{\em solid} arrows through the graph encode members of the set.  For
example, \figref{fig:try_seqbdd}(c) represents the set: $\left\lbrace
ac, abc, aac, acc, abbc \right\rbrace$,where `$acc$' is represented by
$a\rightarrow a \dashrightarrow b\dashrightarrow c\rightarrow
c\rightarrow 1$ (i.e. encode $a$, then do not encode $a$, not $b$,
encode $c$, encode $c$).  Note that a path can include False edges
(dashes), but these do not encode elements of the sequence.  A symbol
in the sequence is generated from the label of the tail of a traversed
edge if the edge is a True edge.




When building a SeqBDD, it is not necessary to first construct a
(potentially huge) binary decision tree and then apply reduction rules
to reduce it. Rather, for SeqBDD, the final compact structure is built
on-the-fly \citep*{Loekito:2010:BDD,denzumi16}, by merging two SeqBDD
graphs, basically composed of node sharing and deletion
rules\footnote{ A faster construction is proposed by use of union
  operator \citep*{Bryant:1986:GAB:6432.6433} \citep*{knuth09}.}.  The
obtained SeqBDD graph is minimal, having the minimal number of nodes
necessary to accept a given set of input.


\subsection{SeqBDD vs. Minimal DFA}

\label{sec:vs}

A SeqBDD is one way to represent a set of sequences in a compact
manner, another is a deterministic
finite state automaton (DFA), defined as follows:  
\begin{quote}
\hspace*{-4mm}acyclic DFA : $\left\langle \Sigma,\varGamma,\delta,s_{start},s_{end}\right\rangle$,  \\
SeqBDD : $\left\langle \Sigma,V,E,s_{start},\lbrace 0,1\rbrace \right\rangle$,  
\end{quote}
where each element in the $\left\langle \ldots \right\rangle$ tuples
denotes the set of accepted elements (the alphabet so far), set of
nodes, set of edges, start\footnote{The start states are suppressed in
  Figures 1 and 2, to save space.}  and end states, from left to right
respectively.
They differ essentially in that BDD labels a node, whereas DFA labels
an edge.  A Trie is one such DFA.  For example, a Trie representing
the set of sequences $\left\lbrace ac, abc, aac, acc, abbc
\right\rbrace$ is shown in \figref{fig:try_seqbdd}(a).

Similar to the original binary decision tree, there are many common edges which
can be shared, such as the edges from `$c$' to the final node.  A minDFA, a DFA
with the smallest number of nodes that accepts the same set of
sequences, can be constructed using the algorithms, originally proposed in
\citep*{Daciuk00}. Briefly, 
two nodes are contracted when the labels of the
incoming edges and outgoing edges are identical. For our
example, the proposed algorithm generates
\figref{fig:try_seqbdd}(b).  (b) is considerably more compact than (a).
We now have two different compact graphical representations: the
SeqBDD (\figref{fig:try_seqbdd}(c)) and the minDFA
(\figref{fig:try_seqbdd}(b)) for the same set of sequences.

Comparing the two graphs even in this simple toy example, we see three
advantages of SeqBDD (c) over minDFA (b).  First, and above all, with
respect to our objective of template extraction, (b) does not show the
template $aXc$, since the $c$ labels are distributed over multiple
edges and are not contracted. This is different in (c), where we
clearly see the template. This natural ability to encode templates
arises from the fundamental difference that the SeqBDD labels {\em
  nodes}.

Second, (b) is less compact compared with the same representation in
the SeqBDD shown in (c), since (b) has 9 labeled graphical objects,
whereas (c) only has 6. This can be theoretically analyzed using the
mathematical notation presented at the beginning of this subsection. 
Let $|X|$ denote the number of elements of a given set $X$. 
It has been recently proven \citep*{denzumi16}, that minDFA requires
$O(|\delta|)$ whereas SeqBDD $O(|V|)$ computational space complexity,
and furthermore, $|V|\leq |\delta|$ where minDFA can be $|\Sigma|$
times larger than the former \citep*{denzumi16}.  In other words, a
SeqBDD is theoretically never larger than the equivalent
minDFA. Moreover, the article reports an empirical comparison of
memory use and SeqBDD reduced memory requirements by 10-22\% relative
to a minDFA.  This is a considerable amount when considering that (as
will be explained in the experimental section) a single target
template can require several gigabytes of allocation.



Third, the algorithm proposed in  \citep*{Daciuk00} is entitled {\em
incremental}, but it is applied to a {\em given graph}
incrementally. In other words, the algorithm of  \citep*{Daciuk00} is not
online. This can be very costly for a large-scale natural language
application, since a huge Trie graph structure must be rendered in
memory prior to applying the algorithm  \citep*{Daciuk00}. This
contraction requires $O(|\delta|)$, computational time complexity
where $\delta$ is the number of edges.  In contrast, as noted earlier,
the application of  \citet*{Loekito:2010:BDD,denzumi16} effects a
direct, incremental construction of the minimal representation.



\section{Relaxed SeqBDD}
\label{sec:relaxed_seqbdd}

\begin{figure} [t]
\centering
\includegraphics[width=0.35\columnwidth]{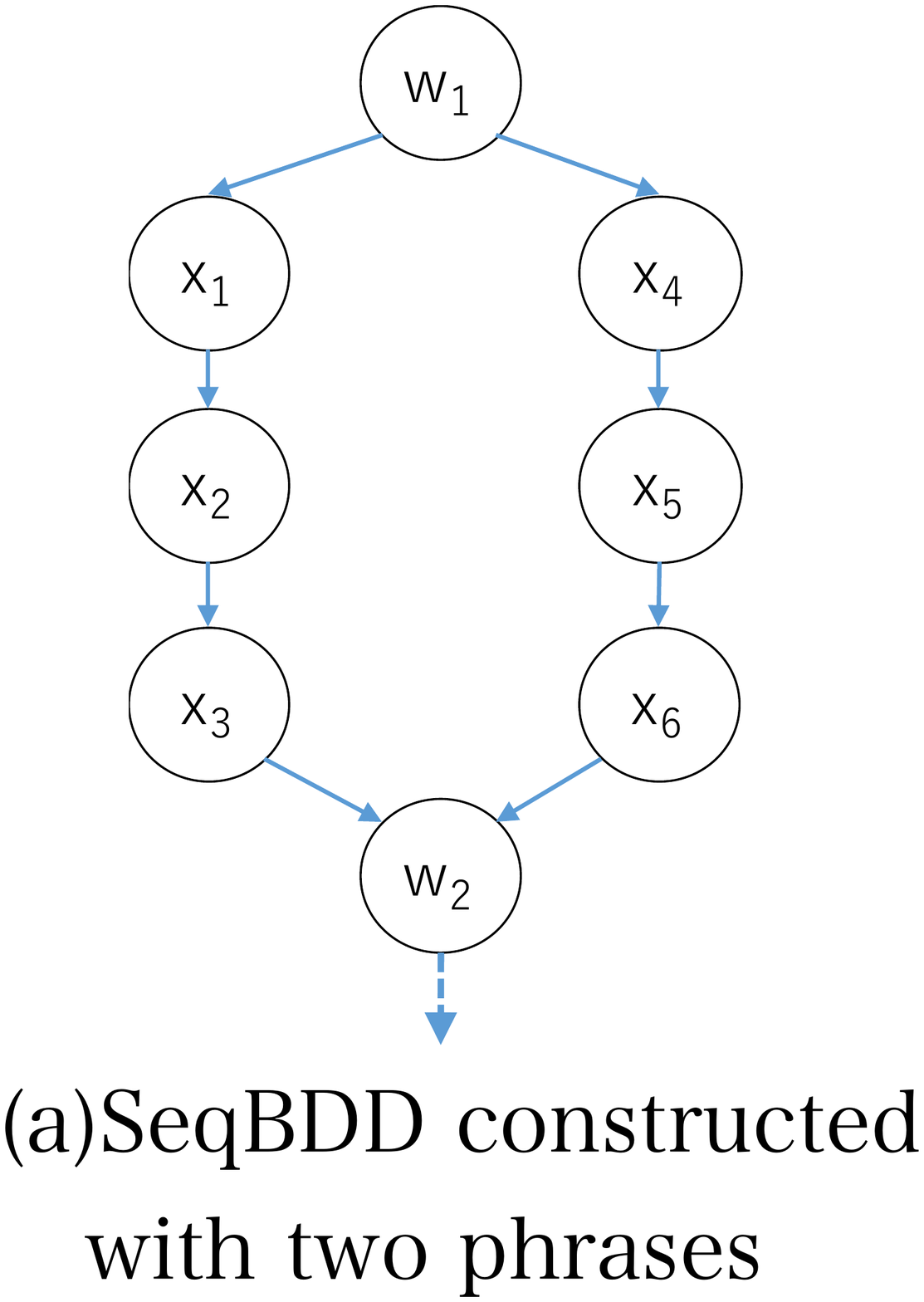}
\hspace*{1cm}
\includegraphics[width=0.35\columnwidth]{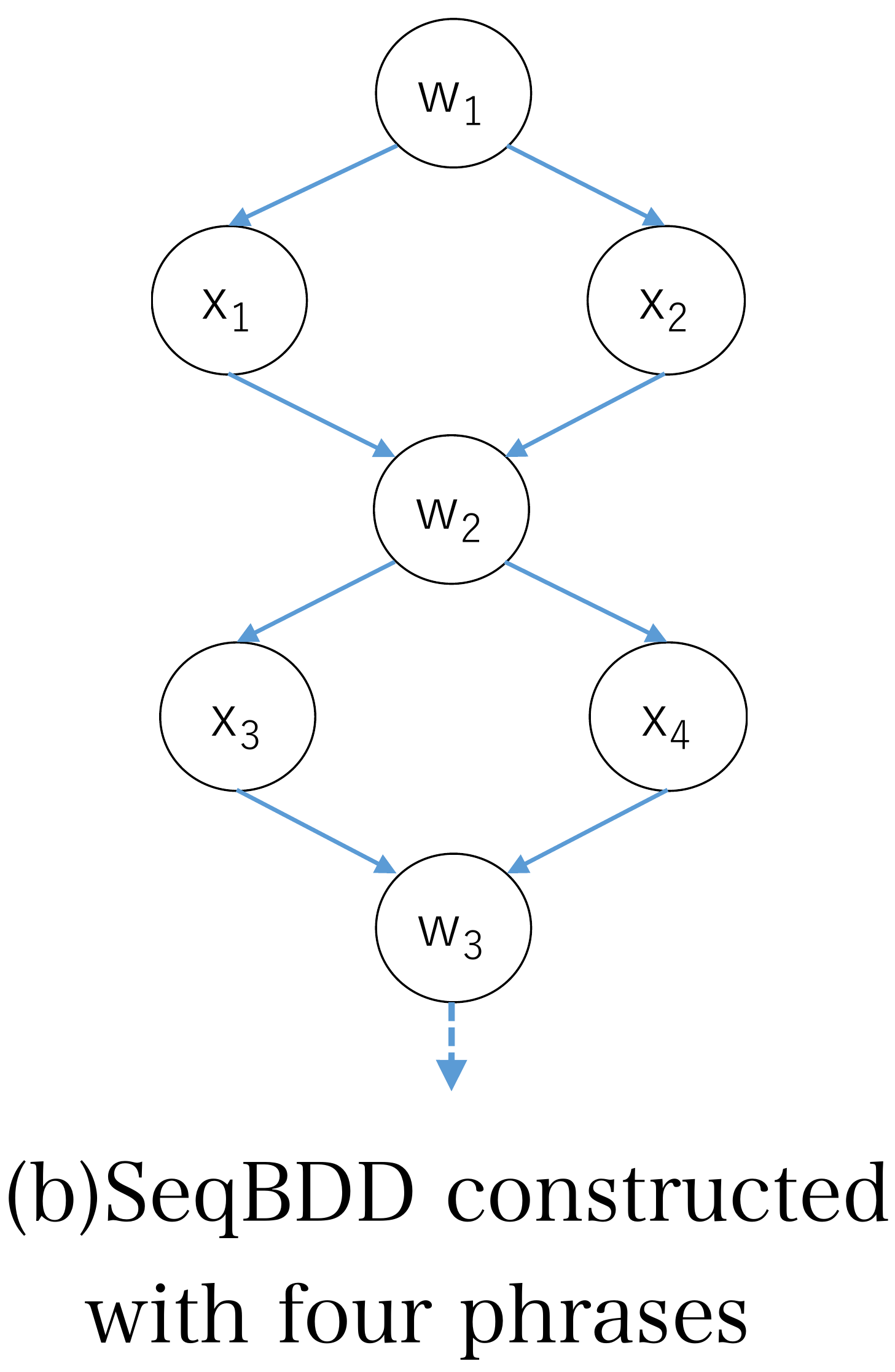}
\caption{
(a) is when two phrases $\{ w_{1}x_{1}x_{2}x_{3}w_{2},
  w_{1}x_{4}x_{5}x_{6}w_{2} \}$ are given as input. (b) is when four phrases 
$\{ w_{1}x_{1}w_{2}x_{3}w_{3}, w_{1}x_{1}w_{2}x_{4}w_{3}, w_{1}x_{2}w_{2}x_{3}w_{3},w_{1}x_{2}w_{2}x_{4}w_{3} \}$ are given as input.}
\label{fig:relaxed_sample}
\end{figure}

As will be shown in the evaluation section, a straightforward
application of SeqBDD as a means to extract a set of templates leads
to a result of low coverage (recall), although the quality of output
templates is high.  This is a consequence of the strictness of the
node sharing operation.  The SeqBDD faithfully represents the supplied
data set, but it is not clear whether this data structure is optimal
with respect to template extraction.


An example is given in \figref{fig:relaxed_sample}, where the
$w_j$ are lexical items and $x_i$ are the potential slots.  The
original SeqBDD can learn the structure in the left figure from just 2
training examples (i.e. $\{ w_{1}x_{1}x_{2}x_{3}w_{2},
w_{1}x_{4}x_{5}x_{6}w_{2} \}$).  To learn the structure shown in the
right figure, however, it requires examples of 4 different forms to
appear (i.e. $\{ w_{1}x_{1}w_{2}x_{3}w_{3}, w_{1}x_{1}w_{2}x_{4}w_{3},
w_{1}x_{2}w_{2}x_{3}w_{3}, w_{1}x_{2}w_{2}x_{4}w_{3} \}$). If even one
of these phrases did not appear in the input set, then the graph
structure gets separated, as in for example the structure in the left
figure and the potential slot before and after $w_2$ cannot be
captured.  In general, when a template has $n$ slots, a number of
corresponding phrases that must be included in the input set grows
combinatorially with respect to the variety of the items around the
$n$ slots.  Although such combination is possible in natural language,
the huge number of required occurrences cannot be actually observed in
real data.

In order to cope with this strictness, we relax the algorithm given by
 \citet{denzumi2011efficient}: we call the proposed graph structure {\bf Relaxed
 SeqBDD} and explain it below.  Since we cannot expect all examples
 to occur in reality, we instead look at {\em False}
phrases represented by the graph. 
The set of False phrases represented by the SeqBDD include
the following two types:
\begin{enumerate}
\item phrases that did not occur
\item phrases that should not occur
\end{enumerate}
In SeqBDD, the 1 is not represented, whereas the latter is omitted: if
represented using a less compact form of original BDD, they are
connected to 0-terminal nodes.
The issue is that possible phrases that did not appear get included in
2 by the standard process of SeqBDD construction.

\begin{figure}[t]
\centering
\includegraphics[width=0.80\columnwidth]{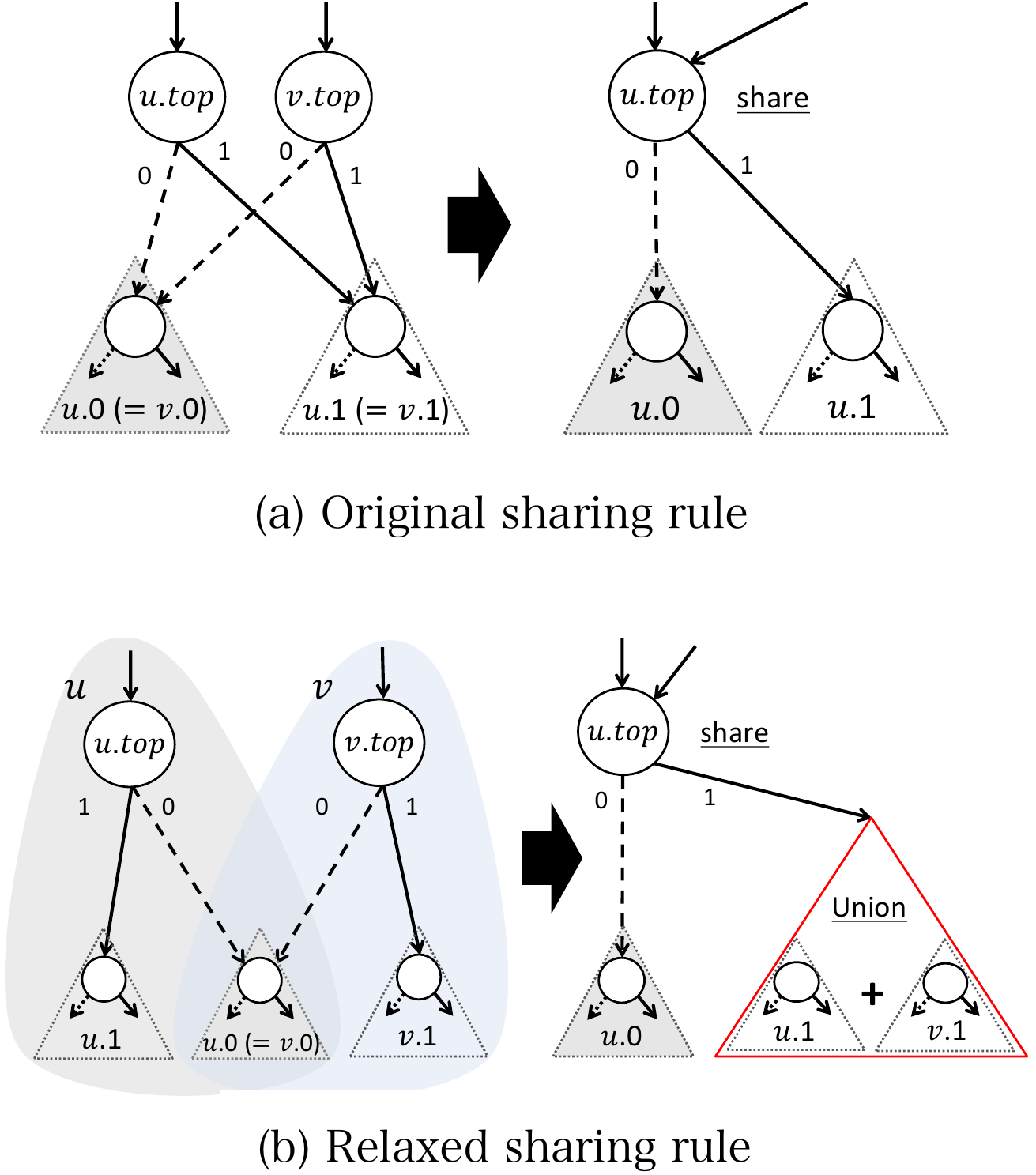}
\caption{Original and Relaxed Sharing rules}
\label{fig:sharing_rule}
\end{figure}

The idea of the Relaxed SeqBDD is to relax the algorithm so that all
the phrases that share 0-terminal nodes get merged during
construction.  To realize this, only the node sharing rule of the
construction algorithm needs to be modified: The node sharing rule for
the SeqBDD is described in \secref{sec:seqbdd_vs_bdd}, and is
intuitively depicted in \figref{fig:sharing_rule}. In the figure of
original algorithm (a), one can see that nodes are only shared when
{\em both} their True and False subgraphs are equivalent, which is
called {\em zero suppression reduction rule}, in the study of BDD.  In
the proposed Relaxed SeqBDD as shown in \figref{fig:sharing_rule}(b),
this rule is replaced by the relaxed version that shares the subgraphs
having the common subgraph leading to the 0-terminal node.  As a means
to satisfy this constraint, the True children are combined using the
union operator proposed in \citep*{Loekito:2010:BDD}, where its
algorithm is provided in Appendix A.  The combination process results
in the sharing of the structures of both True children, and is
illustrated in \figref{fig:sharing_rule}(b); in the figure, the two
True subgraphs of $u$ and $v$ are combined using the union operation.


Unlike the original node sharing rule, there is danger of creating a
cyclic graph with the proposed method (for example, in
\figref{fig:relaxed_sample} right figure, a cyclic graph can get generated if node
$w_3$ was unified with node $w_1$). As a consequence sharing is not
performed if it would result in a cyclic graph.  This new rule allows
the BDD structure to represent word sequences that have not occurred in
the data. Our hypothesis is that this will make the structure more
suitable for template extraction, because observations are 
necessary to make the generalizations leading to templates.

The idea is formally defined as follows as a modification of the
original theory of SeqBDD \citep*{denzumi2011efficient}.  In
order to show clearly where the modifications are, we follow a similar
descriptive form to the algorithms presented in the article.  The
modification of the definition is shown mathematically first and then
we show which part of the algorithm must be modified.


Mathematically, a subgraph $v$  
is characterized by the tuple $\langle v.top, v.0, v.1 \rangle$, where
$v.top$ denotes the top node of the subgraph $v$,
$v.0$ and $v.1$ denote the subgraphs under $v$, 
through 1-edge (called True subgraph) and 0-edge (called False subgraph), respectively.  In the original
 \citep*{denzumi2011efficient}, the subgraphs $u$ and $v$ are shared
in the SeqBDD when:
\begin{quote}
 $\langle v.top, v.0, v.1 \rangle ==  \langle u.top, u.0, u.1 \rangle$,
\end{quote}
where the Relaxed SeqBDD shares $u$ and $v$ when:
\begin{quote}
 $\langle v.top, v.0 \rangle ==  \langle u.top, u.0 \rangle$.
\end{quote}
Given this simple modification, the algorithm of the $Reduce$ procedure
shown in Algorithm 1 (which in fact calls Algorithm 2) can be
modified as in Algorithm 3.

\begin{algorithm}[t]
 \caption{$Reduce$ : SeqBDD Reduction algorithm}
 \label{alg:seqbdd_reduce}
 \begin{algorithmic}[1]
   \REQUIRE  $v$ \COMMENT{acyclic graph}
   \ENSURE The reduced graph
   \IF {$v = 0\mathchar`-\ \OR\ 1\mathchar`- terminal\ node$}
   \RETURN $v$
   \ELSE 
   \RETURN $Get\_node$($v.top,Reduce(v.0),Reduce(v.1)$)
   \ENDIF
 \end{algorithmic}
\end{algorithm}

\begin{algorithm}[t]
 \caption{$Get\_node$ : Original}
 \label{alg:origin_getnode}
 \begin{algorithmic}[1]
   \REQUIRE $v.top, g_{0}, g_{1}$
   \ENSURE Graph processed with node sharing and zero suppression reduction rules
   \IF {$g_{1} = 0\mathchar`- terminal\ node$}
   \RETURN $g_{0}$       \COMMENT{zero suppression reduction rule}
   \ENDIF
   \STATE $u \leftarrow hash\_table (\langle v.top,g_{0},g_{1}\rangle)$
   \IF {$u \neq null$}
   \RETURN $u$  \COMMENT{node sharing rule \secref{sec:seqbdd_vs_bdd}}
   \ENDIF
   \STATE $v.0 \leftarrow g_{0}$
   \STATE $v.1 \leftarrow g_{1}$
   \STATE $hash\_table(\langle v.top,g_{0},g_{1}\rangle) \leftarrow v$   \COMMENT{$hash\_table$ : Hash subgraphs globally}
   \RETURN $v$
 \end{algorithmic}
\end{algorithm}

\begin{algorithm}[t]
 \caption{$Get\_node$ : Proposed Relaxed Version}
 \label{alg:relaxed_getnode}
 \begin{algorithmic}[1]
   \REQUIRE $v.top, g_{0}, g_{1}$
   \ENSURE Graph processed with relaxed node sharing and zero suppression rules
   \IF {$g_{1} = 0\mathchar`- terminal\ node$}
   \RETURN $g_{0}$  \COMMENT{zero suppress reduction rule}
   \ENDIF
   \STATE $u \leftarrow hash\_table (\textcolor{red}{\langle v.top,v.0\rangle})$
   \IF {$u \neq null$}
   \IF {\textcolor{red}{$\lnot\exists$ path from $u.top$ to $v.top$}}

   \STATE \textcolor{red}{$u.1 \leftarrow$ $Union(g_{1},u.1)$}
   \RETURN $u$ \COMMENT{relaxed node sharing rule \secref{sec:relaxed_seqbdd}}
   \ENDIF
   \ENDIF
   \STATE $v.0 \leftarrow g_{0}$
   \STATE $v.1 \leftarrow g_{1}$
   \STATE $hash\_table(\textcolor{red}{\langle v.top,g_{0}\rangle}) \leftarrow v$ \COMMENT{Note: hash key differs from \algref{alg:origin_getnode}}
   \RETURN $v$
 \end{algorithmic}
\end{algorithm}


The sharing of two graphs occurs by providing a subgraph $v$ to
the function {\em Reduce}, as in \algref{alg:seqbdd_reduce}.
The graph reduction process is performed by recursively calling the
$Reduce$ function (line 5). 
The $Reduce$ function uses a $Get\_node$ function
that implements the graph sharing and zero 
suppress reduction rule (see \secref{sec:seqbdd_vs_bdd}). 
To realize the Relaxed BDD, we only have to modify
this $Get\_node$ function, where
\algref{alg:origin_getnode} shows the
original $Get\_node$ algorithm and \algref{alg:relaxed_getnode}, the
proposed algorithm.  The differences are shown in red in
\algref{alg:relaxed_getnode}.

The $Get\_node$ function acquires the top node of the subgraph $v$,
and the two subgraphs that are processed by the $Reduce$ function.
Until line 4, the two algorithms are the same (for processing the zero
suppress reduction rule). Then, the sharing is processed by acquiring
the existing common graph as $u$.  The graph is stored in the
$hash\_table$, that maintains a table of constructed graphs (as
pointers), using node and subgraphs as keys.  The key point of the
difference of the proposed Algorithm 3 from the original Algorithm 2 lies
in this $hash\_table$ function. In the original, the $hash\_table$
stores a structure given the triplet $\langle v.top, v.0, v.1 \rangle$,
whereas the modified method is only given the pair $\langle v.top, v.0\rangle$.

\figref{fig:sharing_rule} (a) and (b) show the difference of the sharing
rules of the original triplet $\langle v.top, v.0, v.1 \rangle$ and
the relaxed pair $\langle v.top, v.0\rangle$.  In the original version
of Algorithm 2, if a common part $u$ already exists, then, $u$ is
returned; if not $g_0$ and $g_1$ are linked as $v.0$ and $v.1$.  In the
modified version of Algorithm 3, the procedure is the same, but in
addition, the cyclic condition is examined (line 6),
and if no cycles exists, unification is conducted (line 8).  
This $Union$ function is a predefined function to
unify two subgraphs which is provided in
\apdref{sec:apdx_union}. The operator is exactly the same as
that defined in \citep*{denzumi2011efficient}.

\begin{figure}[t]
\centering
\includegraphics[width=0.7\columnwidth]{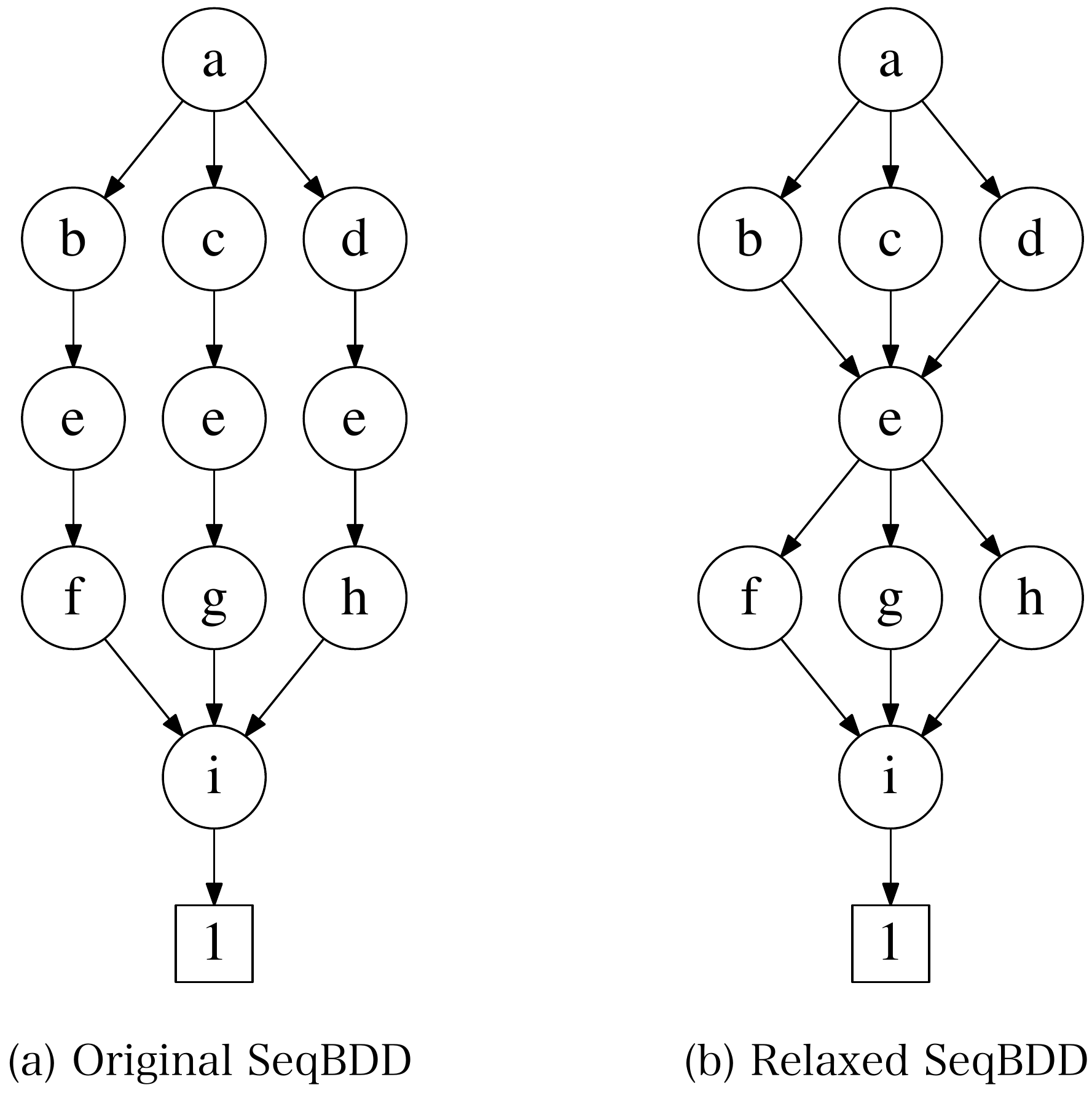}
\caption{Original and Relaxed SeqBDDs  for the input data of $\{abefi,acegi,adehi\}$}
\label{fig:original_and_relaxed_examples}
\end{figure}

In order to show the effect of the Relaxed SeqBDD,
\figref{fig:original_and_relaxed_examples} illustrates two graphs, a
SeqBDD and a Relaxed BDD, generated for the three input sequences
$\{abefi,acegi,adehi\}$.  In (a), the original SeqBDD is shown, where
the $e$ in the middle is not merged. With SeqBDD,
to create the figure as in (b), the input must be $\{abefi,abegi, abehi,
acegi,acefi,acehi, adegi, adefi,adehi\}$, where all combinations of
prefix $\{ab,ac,ad\}$ and suffix $\{fi,gi,hi\}$ should occur.  The SeqBDD
therefore does not incorporate any unseen but possible input.
RelaxedBDD allows the generation of (b), from the three inputs,
assuming the possibility of occurrence of six unseen inputs.  Of
course this is a double-edged sword: the algorithm will certainly decrease
the precision of template extraction of $aXeYi$, but will increase
the recall. The effectiveness of this modification can only be verified
with real evaluation, which we provide in
\secref{sec:multiwordpattern} and \secref{sec:twitter}.
  
\section{Postprocessing of a Relaxed SeqBDD to Obtain templates}
\label{sec:postprocess}



Given a set of word sequences, a Relaxed SeqBDD is constructed by using an
on-the-fly algorithm, thus explained.
Then, we apply the following procedure to obtain the final templates
out of the Relaxed SeqBDD.


The appropriate form of input word sequence depends on the problem
as will be shown in the following two evaluation sections.
In our work, the input word sequences are labeled with POS tags, and the
SeqBDD is constructed based on the POS tags ($\Sigma$ is the set
of POS tags).  The words are nevertheless preserved in the nodes as
auxiliary information, in the form of lists containing the words at
nodes that were contracted through node sharing operations during BDD
construction.  Let $\mathcal{W}_u$ be the set of words at node $u \in
V$, $f(w)$ be the relative frequency of $w \in \mathcal{W}_u$.

\begin{figure}[t]
\centering
\includegraphics[width=0.4\columnwidth]{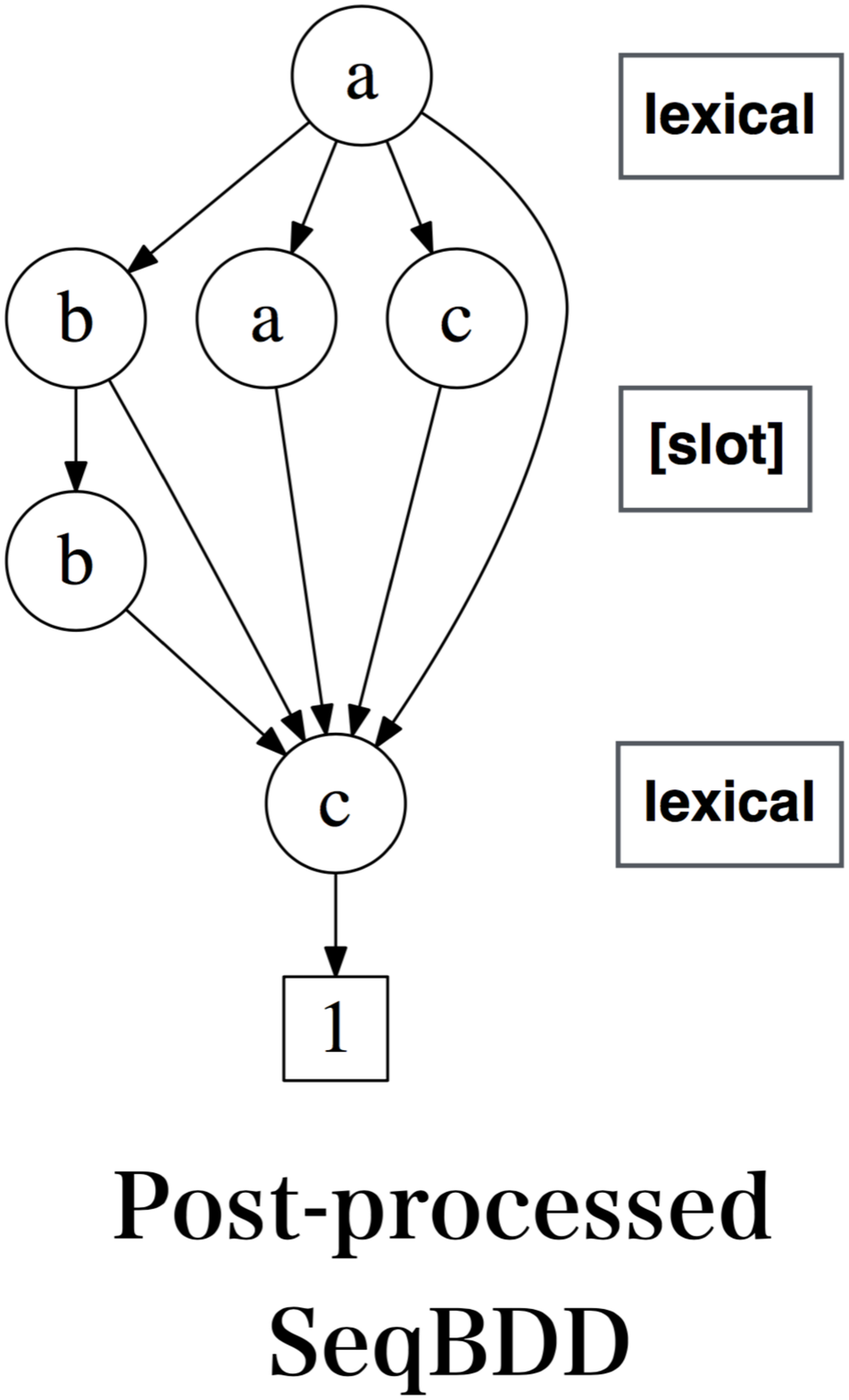}
\caption{Post-processed SeqBDD representing: $\left\lbrace ac, abc, aac, acc, abbc \right\rbrace$ same as \figref{fig:try_seqbdd}.}
\label{fig:post_seqbdd}
\end{figure}

\begin{figure}[t]
\centering
\includegraphics[width=1.05\columnwidth]{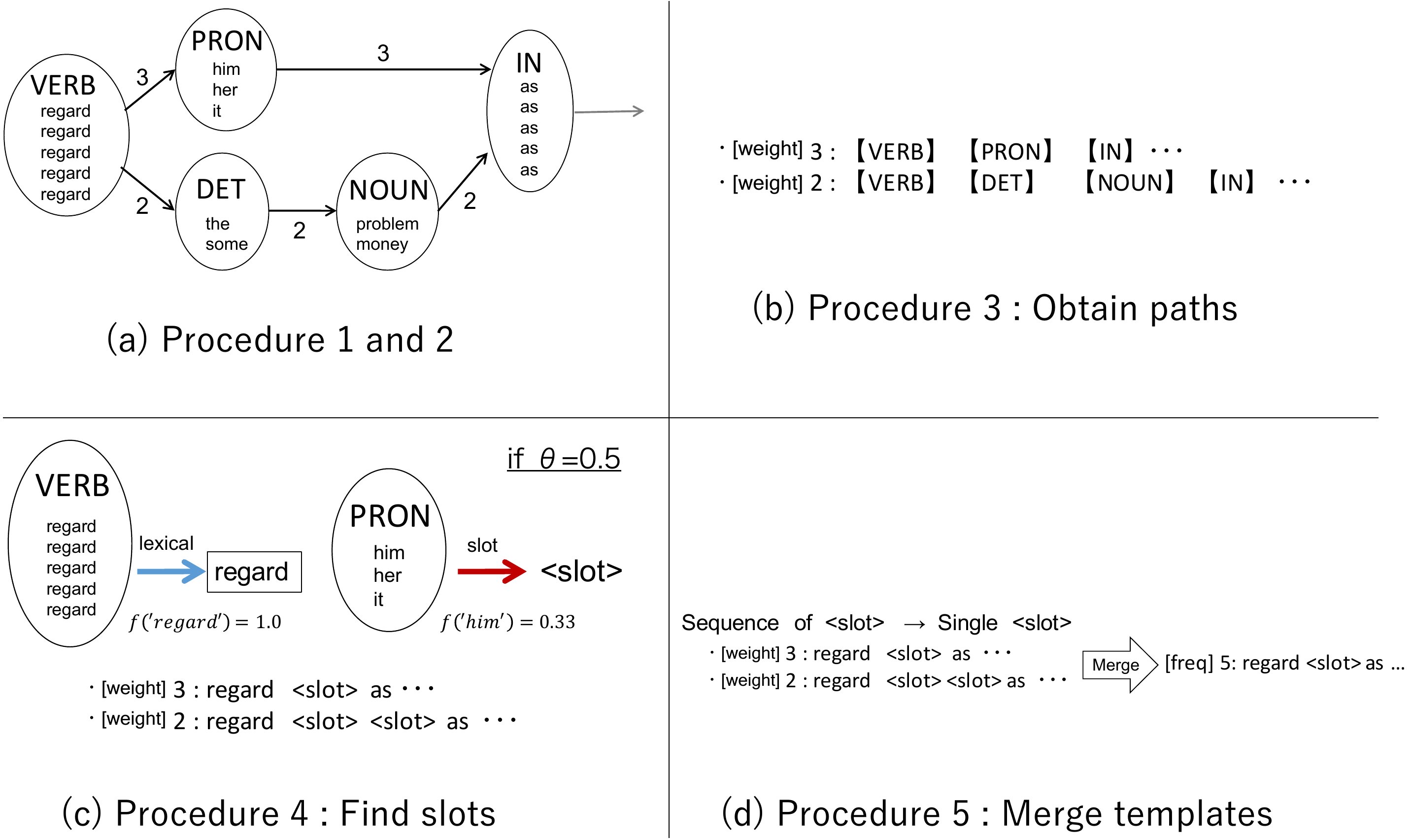}
\caption{SeqBDDs post-processing examples}
\label{fig:post_examples}
\end{figure}

Fully constructed SeqBDDs are post-processed as follows.
\begin{description}
\item[0. POS-tagging:] To build a relaxed SeqBDD, all word sequences must be POS tagged. 
\item[1. Delete Falsehood:] The 0-edges 
are simply deleted, since they are no longer necessary.
\item[2. Weight edges:] The edges are then weighted by the absolute frequency
of the number of phrases that goes through the edge. In order to reduce the complexity of
the graph, all edges with frequency one are removed.(\figref{fig:post_examples}(a))
\item[3. Obtain paths:] Obtain all {\em paths} connecting to the
  starting status of $s_{start}$\footnote{
There are nodes that are not connected to $s_{start}$, caused by elimination of branches at step 2.
}. 
Weight each path with the {\em minimum} frequency of the edge included
in the path. (\figref{fig:post_examples}(b))
\item[4. Find slots:] Each node $u$ along a path is labeled as
``lexical'' if $\max_{w \in \mathcal{W}_u} f(w)  \geq \theta$, ``slot'' otherwise, where
$\theta$ is set to best suit the problem. This
transforms a path into a template with slots. (\figref{fig:post_examples}(c))
\item[5. Merge templates:] 
Merge identical templates and weight the final template
by the sum of the weights of the merged templates. (\figref{fig:post_examples}(d))
\end{description}

The output consists of a frequency-ranked list of {\em templates}.
For example, \figref{fig:try_seqbdd}(c) is transformed into \figref{fig:post_seqbdd}
through procedures {\bf 1} and {\bf 2} above, where the template
`$aXc$' is now readily visible: the output nodes and edges of the real
examples are labeled by the weights.  Procedure {\bf 4} defines which
nodes are slots. 

The overall procedure above is equivalent to determining the paths by
maximal flow  \citep*{cormen}.  Since the edges $e \in E$ of the SeqBDD
are weighed by frequency, the maximal flow in every path is naturally
limited by the minimum frequency edge (Procedure {\bf 3}).  We believe that
this is a natural procedure to extract a template given a SeqBDD
graph, but other possible strategies exist and remain future research.

We show how this postprocessing proceeds using
\figref{fig:post_examples}.  Let us assume that there are 5 input
phrases\footnote{Many NPs in this slot could be longer than this,
  but here short toy examples are given only for explanation.} :
\begin{quote}
 {\scriptsize
\noindent   regard him as $\cdots$\\
   regard her as $\cdots$\\ 
   regard it as $\cdots$\\
   regard the problem as $\cdots$\\
   regard some money as $\cdots$
 }
\end{quote}
The Relaxed BDD is constructed and by eliminating unnecessary
information through procedures {\bf 1} and {\bf 2}, and the graph would look
as that in (a): the graphs are aligned horizontally here (due to space limitations),
nodes indicate shared POS, and edges are annotated with frequencies.
Examination of paths through procedure {\bf 3} gives potential
template candidates as in (b). Then lexical/slot nodes are judged with
the relative frequency function $f$ as in (c) through procedure
{\bf 4}. Lastly, the equivalent templates are merged to obtain the
the final results through {\bf 5} as in (d).



\label{sec:expt_data}

\section{Verb+Preposition Template Extraction}
\label{sec:multiwordpattern}

We first present a large-scale experiment that attempts to extract verb
templates out of given sentences.
A verb+preposition template here is a template
that involves a verb, such as ``regard X as Y''.
This test intends to evaluate the performance of the
method using a clean dataset, acquired from a dictionary.

For this purpose, the verb+preposition templates were collected by
searching Oxford Advanced Learner's Dictionary for phrasal verbs with
slots. This extraction was based on the annotation provided by the
Dictionary, for example ``regard X as Y'' where X and Y are slots.  We
considered verbs with multiple slots, only, and excluded verbs that
only have one slot. The slots can be located at the end of the
template as in ``regard X as Y''.  To collect such instances, a verb
list was acquired from the British national corpus, we scanned the
Learner's dictionary for every verb, and extracted the
verb+preposition templates with slots using regular expressions.  We
obtained a set of 976 templates.




For each template, the sentences containing all words in the template
were obtained from the English Gigaword Corpus (fifth edition;
LDC2011T07), consisting of 180 million sentences lemmatized and POS
tagged using the Stanford CoreNLP tools~
 \citep*{manning-EtAl:2014:P14-5}.  The sentences that included the verb
 were obtained and then filtered using regular expressions so that 
a span between the slots was at most five words.  For example, for the
template ``regard X as Y'', all sentences with the words ``regard''
and ``as'' were extracted, under the constraint that the number of
words between ``regard'' and ``as'' was at most five.  Similar
constraints have often been used in previous work, notably
 \citep*{baldwin-handbook10}. The proposed method and the parser-based
method that we compare with (defined in the following) were then used
to discover the templates from these {\em search phrases}.


For the proposed method, we extracted ranked lists of template
hypotheses from the search phrases, using the method described in
\secref{sec:postprocess}, after the sentences in the search phrases were
reduced to sets of subsequences following the main verb.  Since there is
no established baseline for this task, another plausible method using a
parser was created for comparison.  This method using a parser is an
unfair baseline, since the parser-based method uses
vastly more resources, requiring a parser trained on a structurally
annotated corpus, whereas our method uses only tagged corpus. In fact,
these are two different methods which could even complement each other
for the task of phrase structure induction. Hereafter, we call this
alternative method to be compared with our method, the parser-based method.

\begin{figure}[t]
\centering
\includegraphics[width=0.9\columnwidth]{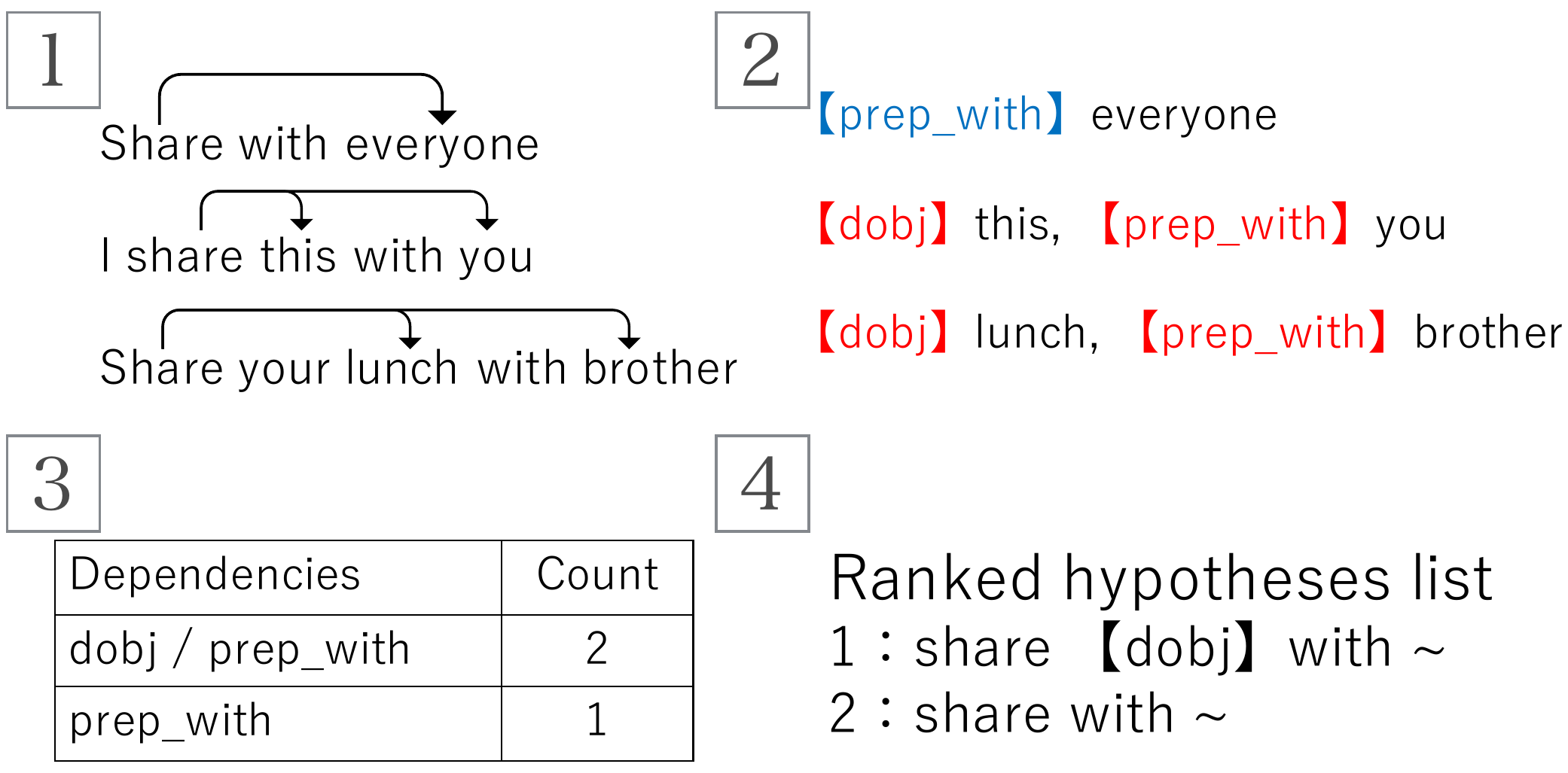}
\caption{Extracting parser-based template hypotheses.}
\label{fig:baseline}
\end{figure}

In this parser-based method, first the search phrases for all of
the 976 templates being studied were dependency parsed. The
parser was the Stanford Parser, based on universal dependencies.
The procedure is conducted by providing the whole sentence to the parser. 
\figref{fig:baseline}-1 shows 3 such sentences for the template ``share
X with Y''. \figref{fig:baseline}-2 shows the dependencies of the
verb in the template. In \figref{fig:baseline}-3 template hypotheses for
the parser-based method are created from sentences that contain the same
dependencies, for example the second two sentences in the figure both
contain the dependencies: \texttt{[dobj]} and \texttt{[prep\_with]} and
these give rise to a single template hypothesis with a count of 2. The
hypothesis counts reflect the number of sentences that share the same
dependencies, and are used to rank the template hypotheses
(\figref{fig:baseline}-4).




The performance was evaluated by mean reciprocal rank (MRR)
 \citep*{Manning:2008:IIR:1394399} and recall, to judge the quality of the lists of
extracted template hypotheses.  Recall is the proportion of the correct verb
template that were extracted.
As for what corresponds to the precision, 
since the output templates are ranked and the rank must be evaluated,
we used MRR. MRR is the average of reciprocal rank of the
correct verb+preposition template in the output list.
Let $Q$ be the number of
test cases, for each of which the procedure extracted a template at rank $r_i$ ($i=1,\ldots,Q$).
\begin{equation}
  MRR = \frac{1}{Q} \sum_{i=1}^Q \frac{1}{r_i}
\end{equation}
When the correct template was not extracted for a set of search phrases, 
its rank was considered to be $\infty$.
In this experiment, $Q=976$.

\begin{figure}[t]
\centering
\includegraphics[width=0.7\columnwidth]{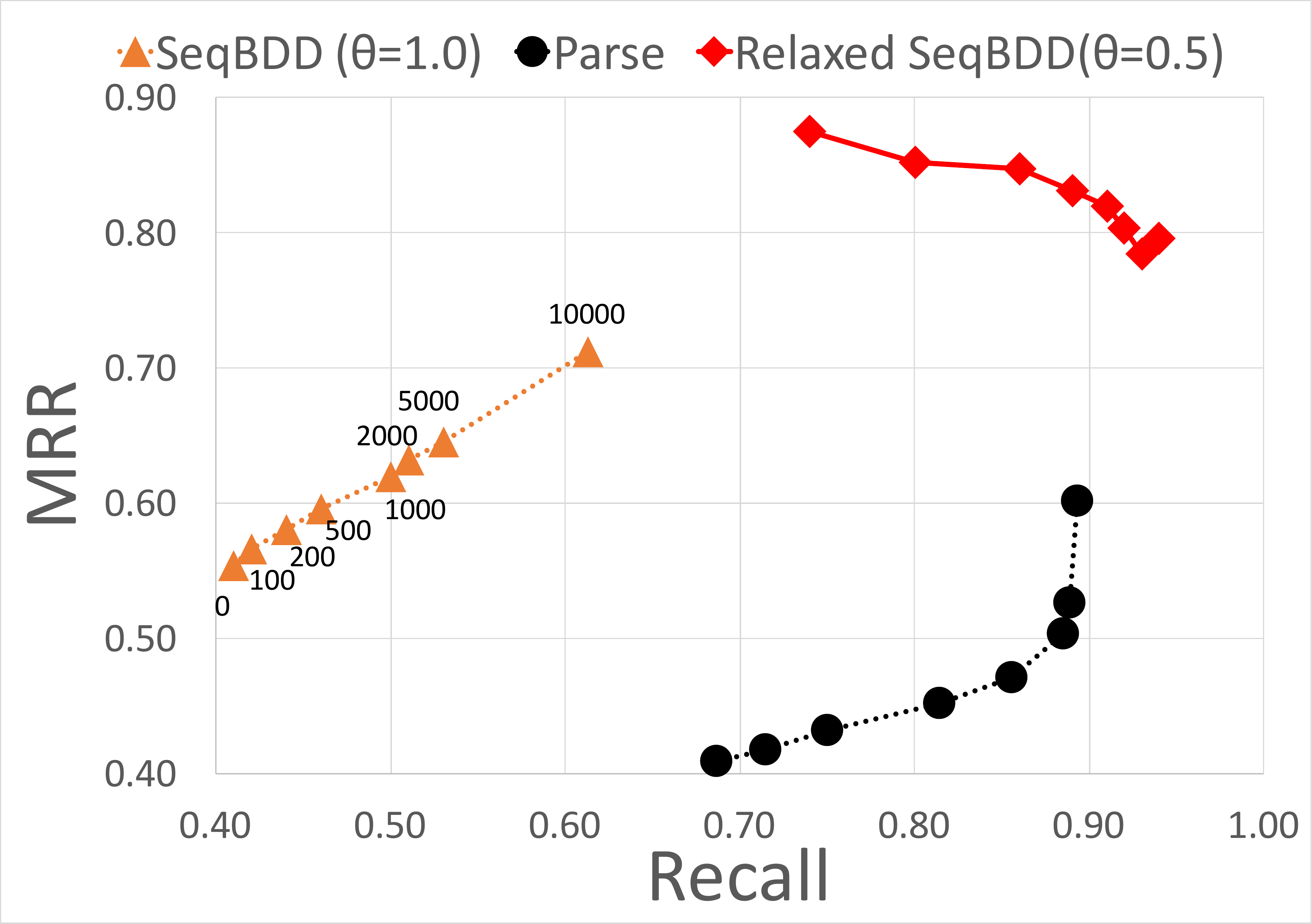}
\caption{verb+preposition template extraction from the Gigaword corpus}
\label{fig:map_result}
\end{figure}

The results are shown in
\figref{fig:map_result}.  To observe the effect of dataset size,
dataplots show the average MRR and recall of templates that had {\em
  more than} a specific cutoff threshold number of search phrases.
The cutoff values used were: 0, 100, 200, 300, 500, 1k, 2k, 5k, and
10k. These values are annotated only on the `SeqBDD($\theta=1.0$)'
orange curve on the graph; on all other curves the cutoff variables
run in the same manner.

We first examine the parser-based method and the two variants of the proposed method as follows: 
\begin{description}
\item [Parse] in black dotted line. The parser-based method.
\item [SeqBDD ($\theta=1.0$)] in orange dotted line. (For $\theta$, see \secref{sec:postprocess}.)
\item [Relaxed SeqBDD ($\theta=0.5$)] in red line with triangles. 
\end{description}

%


It can be seen in \figref{fig:map_result} that the SeqBDD method is
able to achieve higher levels of MRR than the parse-based baseline,
but at a cost in terms of its recall.  The proposed Relaxed SeqBDD
variant is able to achieve considerably higher levels of MRR in all
cases than both methods without losing ground in terms of recall.
It is impressive how higher MRR was acquired: this is
due to the relaxed algorithm that integrates template candidates scattered
in the original SeqBDD.  We observed that for the Relaxed SeqBDD there is a
trade-off between MRR and Recall, however, the Relaxed SeqBDD was able
nonetheless able to achieve MRR levels greater than the other methods
even when the level of recall was at its maximum value of 95\%.  It is
clear from the figure that the baseline and SeqBDD methods depend
heavily on the amount of data, whereas the proposed method appears to
be far less sensitive to the data set size, and is able to achieve
respectable MRR and recall on even the smallest subset of the data
used in our experiments.  Note, that the parser-based method used a
parser trained on a structurally annotated corpus, whereas our method
used only a tagged corpus.

The end-to-end runtime performance\footnote{In this work, we used the Python language
on a PC cluster with CPU of Intel E5 3.4GHz and 512GB of RAM running
on CentOS.}, from raw text until the final output for the proposed SeqBDD
method was 72.1 sec and 866.7 MB per template. The Relaxed SeqBDD method required 
247.1 sec for processing and 1.5GB of memory. The baseline was the fastest, taking
only 51.8 sec for processing, but required 2.3 GB of memory due to the resources
required for parsing.

\section{Template Extraction from Twitter}
\label{sec:twitter}

In this section, we are
concerned with a more real-world application: 
the extraction of templates from short text
messaging. Twitter, for example, is used for commercial objectives, and
it is relatively common for the information in tweets to be generated using templates, as in the
following examples:
\begin{quote}
{\scriptsize
\noindent
5.1 earthquake, Kermadec Islands region. 2016-06-23 07:44:37 UTC at epicenter (24m ago, depth 55km).\\
5.5 earthquake, Scotia Sea. 2016-06-23 00:05:39 at epicenter (20m ago, depth 10km).\\ 
5.1 earthquake, Southern Mid-Atlantic Ridge. 2016-06-21 17:47:36 at epicenter (19m ago, depth 10km). \\
6.1 earthquake, Northern Mid-Atlantic Ridge. 2016-06-21 13:26:35 at epicenter (31m ago, depth 10km).\\
5.3 earthquake, South of the Fiji Islands. 2016-06-21 23:40:40 at epicenter (24m ago, depth 565km).
}
\end{quote}
The automatic extraction of such templates will 
facilitate the analysis and extraction of structured information. 
The task in this experiment is to output the templates for a given set of tweets.
Clearly this example is output by a machine bot: other more real examples
are provided in Appendix B.


For this purpose, we collected corpora of Tweets in English and
Japanese.

In the case of English, a set of Twitter accounts was mined starting
from a single Twitter account by recursively following the followers
of the accounts. A set of tweets was collected for each account
traversed. In the case of Japanese, a list of tweet sets for accounts
was downloaded directly from the
web\footnote{URL:http://seesaawiki.jp/w/wikkkiiii/}.
Since we are
interested in evaluating template extraction, we downloaded the ones
which were judged to contain templates by the baseline procedure.
Extraction from random data is unlikely to be fruitful, since any extraction
algorithm will not work unless many similar expressions are contained in
the original data.  Therefore, we have to filter the dataset first.

The baseline procedure is based on heuristics of substring matching,
described as follows.  For a set of tweets for a Twitter account,
$k$-means clustering \citep*{Macqueen67somemethods} was applied with
the similarity between two tweets being the word TF-IDF vector
\citep*{Deerwester90indexingby}.  The procedure partitions the set of
tweets into $K$ clusters.  Templates were constructed for each cluster
as follows. First, the common words that appear in more than half of
the tweets were identified and preserved as words in the template;
then single slots were created to replace the sequences of remaining
words.  The most frequent template was output for each cluster,
leading to $K$ templates being output for each set.  Here, we set the
value of the $K$ to 20, that presented the highest F-score in our
final evaluation.  If any of the templates contained one slot between
two lexical items, then the list of tweets for the set was said to
contain a template.  The length of the expression filling the slot
could be longer than five words, different from the previous evaluation
task of verb+preposition template.

This baseline was used to produce the baseline output, and also
 to create the gold standard explained as follows. 
First, using this baseline, we collected a set of 600 tweet sets (1 set per Twitter account) of
English tweets, and another set of 600 tweet sets of Japanese
tweets: for every set, the template suggested by the baseline contained more than one slot,
and we excluded a set that had short templates with only one slot.
Then, trivial noisy tweets, such as replies and re-tweets
together with those with URLs in the tweets were deleted or replaced
automatically by using regular expressions.  They are able to form
(unwanted noise) templates, and they can affect performance, depending
on their frequency of occurrence in the tweets.

After cleaning, we constructed two types of gold standard from this preprocessed data.
The two annotators\footnote{ 
  One annotator, who is unrelated to this project, was hired specifically for the annotation task. 
  The other annotator is the first author.
  The two annotators initially worked separately.
  Given the baseline, the annotators were able to
  observe the appearance of templates existing in each of the datasets
  and they worked by modifying the appearance and borders of the templates.
  The agreement of the two annotators was pretty high, due to the
  baseline highlights, even though they were often implausible. 
  Then the results were discussed by the two annotators to reach a complete agreement,
  and the annotators rectified their errors and omissions.
  The overall result was checked by the second author.
  This procedure was conducted before the
  actual experiments of SeqBDD, without knowing its output. 
} manually 
scanned every tweet set, in which the candidate template is highlighted
by the baseline method. 
First, every Twitter set is annotated with whether it contained a template or not.
Second, among those which contained a template,
we randomly chose 100 sets and annotated them with the templates; each account was
annotated with multiple possible templates, that were
found in many tweets included in the account.

This set of 100 sets is defined
as {\bf Template-annotated corpora}, and the rest is defined as {\bf
Hand-classified corpora}.  Note that a set of the former always
contains a template, whereas an account of the latter does not
necessarily have one. Relatedly, since the baseline procedure is used
to collect the data, the recall of the baseline system is 1.0.

Each set of tweets from the template-annotated corpora was annotated with a maximum of five
templates, created by an independent human annotator (not one of the
authors).  For example, the set of tweets in the example above was annotated with the template:
\begin{quote}
\textcolor{red}{\scriptsize $<$slot$>$ earthquake , $<$slot$>$ . $<$slot$>$ at epicenter ( $<$slot$>$ m ago , depth $<$slot$>$ km ) .}
\end{quote}
A summary of the template-annotated and hand-classified data is shown
in \tabref{tab:tweets_data_sets}.  Typical sources included news,
weather and public information feeds.  A set of tweets consisted of
all tweets from a single source. The total size of the
corpora was over three million tweets.

It can be argued that the dataset is biased in favor of the baseline
since it was created using the baseline.  Without use of a tool that
gives a hint to human annotators wherein a tweet set contains a
template, however, the annotation task was too difficult and time
consuming. Since we cannot use the proposed method for this task, we
used the baseline, even though the resulting comparison will be in
favor of the baseline.  However, as will be shown, the proposed method
outperforms the baseline even on such data. Therefore, we believe that
the evaluation serves to demonstrate the effectiveness of our
approach.


\renewcommand{\arraystretch}{1.3}
\begin{table} [t]
\centering
\caption{Summary information and statistics for the Twitter data-sets.}
\label{tab:tweets_data_sets}. 
\small
\begin{tabular}{lll}\hline
Tweets-set  & {\bf Template-Annotated}    & {\bf Hand-classified} \\ \hline
Languages and & each 100 English    & each 500 English\\
the number of sets & and Japanese   &  and Japanese \\
&&(39 English and 23 Japanese \\
&&do not have templates) \\ \
Label   & Manually annotated templates & Manually classified whether  \\
       & (a maximum of 5)        & a template exist or not \\ 
The number of & from 478 to 3200    & from 261 to 3200 \\ 
tweets range/Ave.& /2578 tweets per sets & /2576 tweets per sets \\
Experiment & Both Pattern Classification    & Only Pattern Classification \\
           & and Extraction&  \\\hline
\end{tabular}
\end{table}
\renewcommand{\arraystretch}{1.0}


Two types of experiment were performed. The first used all the  600 data both
in English and Japanese.  We call this experiment the {\bf Pattern
  Classification Experiment}.
The task is a straightforward classification task to identify whether
or not templates exist in the tweets, and we therefore used the
standard evaluation method of precision/recall/f-score to evaluate the
performance. The results are shown in
\tabref{tab:template_judge_result}.

The precision of the baseline is 0.96 for Japanese and 0.93 for
English, indicating that the heuristic used to extract the candidate
template data was of high quality. Nonetheless, applying the proposed
Relaxed SeqBDD method to this data was able to give respectable gains
in both precision and f-score at a small cost in terms of recall. It
is also clear from the table that Relaxed BDD is superior to
the method based on the original SeqBDD in all respects.

\begin{table} [t]
\centering
\caption{Template extraction results for all the 600 set of data.}
\label{tab:template_judge_result}
\begin{tabular}{lllll}\hline
&  & {Baseline} & SeqBDD & Relaxed SeqBDD \\ \hline
& P & 0.964 & 0.978 & 0.983 \\
Japanese & R & 1 & 0.816 & 0.985 \\
& F & 0.982 & 0.889 & 0.984 \\
&&&&\\
& P & 0.935 & 0.954 & 0.975 \\
English & R & 1 & 0.836 & 0.978 \\
& F & 0.966 & 0.891 & 0.976 \\\hline
\end{tabular}
\end{table}

The second experiment evaluated the quality of the templates that were
actually extracted, using the Template-annotated corpora (100 each in
English and Japanese).  We call this experiment the {\bf Pattern
  Extraction Experiment}.  Since the baseline outputs $K$ templates,
only the best $K$ templates were considered also for SeqBDD methods.
Similar to the previous work in \figref{fig:map_result}, these ranked
lists were evaluated by mean reciprocal rank (MRR) of the best ranked
correct template appeared in the output list of length $K$.
Here, in formula (1), $Q=200$.
If there are multiple correct templates
extracted, the reciprocal of the best rank is used to obtain the
average, and if no correct template was found, then the rank is set as
$\infty$.

\begin{figure}[t]
\centering
\includegraphics[width=0.7\columnwidth]{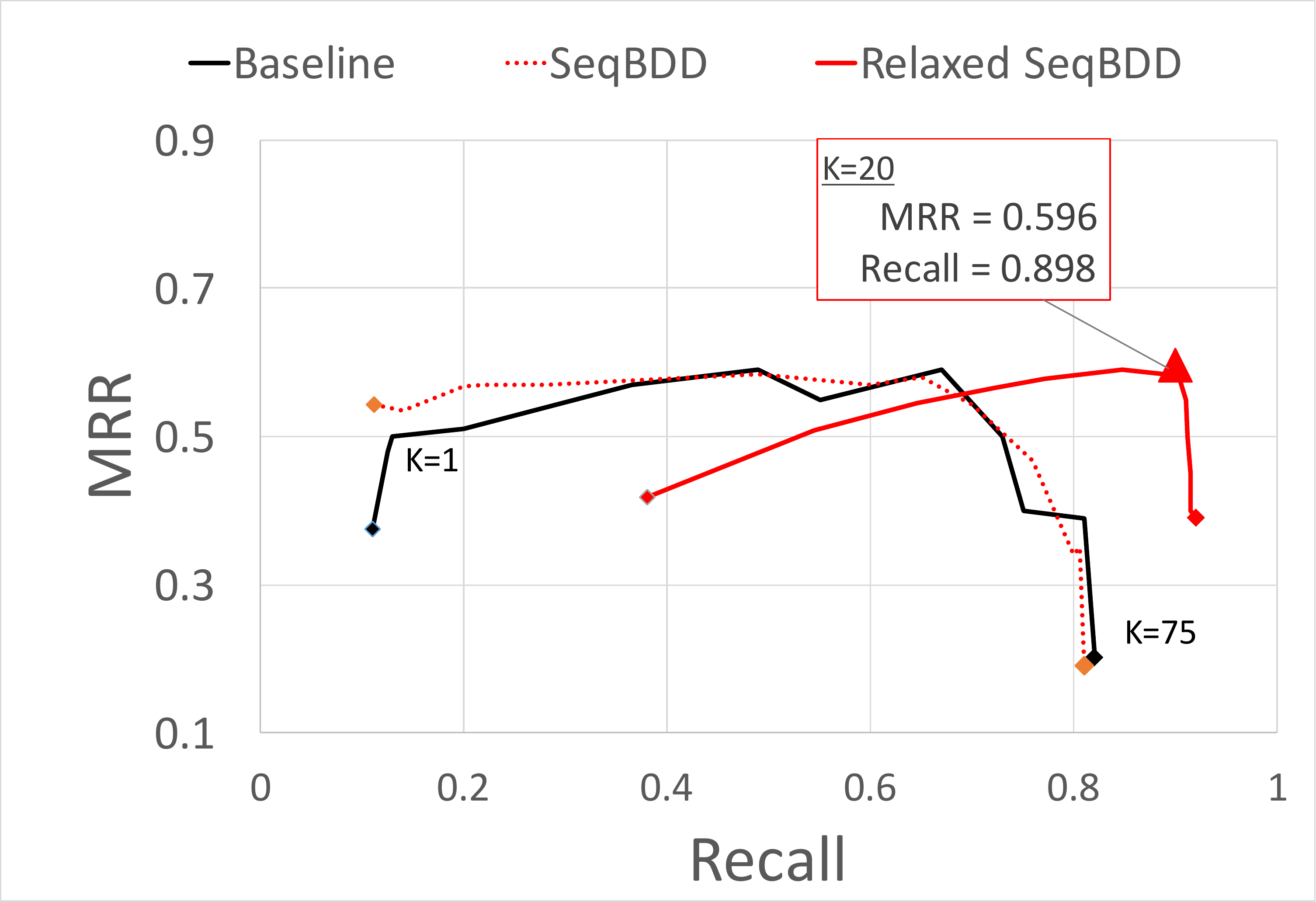}
\caption{ Extraction of Templates from Twitter using Template-annotated corpora.}
\label{fig:template_extraction_result}
\end{figure}

The results are shown in \figref{fig:template_extraction_result},
English and Japanese altogether.  There are three lines: black, orange
dashed and red solid, representing the baseline, SeqBDD and Relaxed
SeqBDD respectively.  Each line shows the values of MRR and $R$, while
varying $K$.  In all cases the recall of the Relaxed SeqBDD was
considerably higher that the SeqBDD method, this was expected and was
caused directly by the relaxation of the algorithm allowing it to be
applicable to more cases in the data. There is of course a trade-off
with MRR, although this trade-off proved to not to be too detrimental,
and in fact for all values of $K$ greater than 3, the Relaxed SeqBDD
achieves higher levels of both MRR and recall than the other methods.

The end-to-end runtime performance,
from raw text until the final output for the proposed SeqBDD
method was 37.3 sec using 836 MB of memory for a single tweet-set, the Relaxed SeqBDD method took 210.5 sec
and used 1.4GB, whereas the baseline took 30.9 sec and used 2.3 GB.

Some example extraction results for the Twitter data are shown in the
Appendix B.  The patterns are successfully acquired for Twitter
examples like B.1 or B.2, where the pattern frame structure is solid,
and the length of the expression filling the slot is short.

On the other hand, the algorithm fails to extract the correct patterns
when the length of expression filling the slot is large, such as in
the example of B.4, due to the fact that the Sequence BDD graph
becomes too sparse.  For example, patterns such as ``How to * (URL)''
or ``Outsourcing Human Resouces * (URL)'' could have very long
expression in *.  Moreover, if the preprocessing of POS tag fails,
then all that follows also fails, as shown in the example of B.3.  The
development of methods for overcoming these issues remains future
work.

\section{Conclusion}

This paper presents a novel method, based on an original relaxed
algorithm for building SeqBDDs, for inducing a template ---a subsequent
phrase with slots---, given a set of phrases having common structure.
The SeqBDD whose construction process we relax, is equivalent to a
minDFA but more efficient and well-suited for template extraction. We
proposed a relaxed form of the SeqBDD as a means to represent sets
of word sequences graphically in a compressed form that naturally
discovers structures in the text. The proposed enhancement to the
SeqBDD construction process, relaxes the node sharing rule to
facilitate a greater degree of node sharing. The structures inherent
in the SeqBDD constructed from the text are then extracted to produce
templates.

We evaluated the proposed relaxed method on two template extraction
tasks: verb+preposition templates, and tweet templates.  In verb+preposition template
extraction, the original SeqBDD was able to achieve considerably higher
MRR than a parser-based method, but had lower recall.  When relaxing
the SeqBDD, the method was able to substantially exceed the
performance of parser-based method both in terms of MRR and recall.
In the more real-world task of tweet template extraction using more
than three million tweets, our experiments show that the proposed
relaxed SeqBDD method was able to outperform the original SeqBDD by
substantially increasing recall while at the same time maintaining
high levels of precision.



This work represents only a first step on the road to inducing
templates. This paper has shown empirically that relaxing the SeqBDD
to coalesce very similar, but slightly different, structures greatly
enhanced its utility for template extraction. We believe further
study along these lines would be fruitful in the future. In addition,
we intend to focus on improving the efficiency of the algorithm in
future work to allow it to scale for use on large data sets. This
requires a further approximation of the method, and 
should be evaluated from the point of view of both accuracy and speed. 

In the future, our work could be linguistically extended from the
viewpoint of the Pattern Grammar, our original motivation; the current
work was limited by being simplified in order to be realized as a
computational procedure. 
Since the Pattern Grammar has been supplemented recently by work on
Construction Grammars from Cognitive Linguistics, one possible future
path might be to reformulate the output templates in relation to more
formal linguistic notions. 

Finally, as for the downstream applications of our method, an
implementation of KWIC that produces templates would be helpful for
lexicographers and corpus linguists. Moreover, we believe that a module
for SNS and blog data analysis would prove useful from industrial
perspective.


%

\bibliographystyle{acmtrans}
\bibliography{nle}

\include{appendix}

\appendix

\nopagebreak
\section*{}
  \label{lastpage}

\end{document}

%% file: appendix.tex
\appendix
\section{Union operation of SeqBDD}
\label{sec:apdx_union}

This section explains the Union operator  
as proposed in \citep*{Loekito:2010:BDD}. 
This Union operator is used in Algorithm 3 line 7. 

Given two SeqBDDs $P,Q$, The operation to produce 
$P\cup Q$ is shown in \algref{alg:union}.

First, if either of $P$ and $Q$ are 0-terminal nodes,
then the one which is not the 0-terminal node is output.
If $P$ and $Q$ are equivalent, then $P$ is output.

In other cases, the top nodes of the two graphs are compared.
If they are the same, then the new top is the $P.top$
and the true/false subgraphs are constructed by 
merging $P$ and $Q$'s true/false subgraphs respectively,
where the merging is conducted using the Union operator. 

Suppose that $P.top$ is alphabetically smaller than $Q.top$, then
$P.top$ is smaller than any nodes represented in $Q$.  Therefore, the
new top is assigned to be $P.top$, and $P$'s true subgraph 
ends with the 1-terminal node.  $P$'s false subgraph must be merged
with $Q$ where this operation is conducted recursively using the Union
operator.



\begin{algorithm}[ht]
 \caption{$Union$}
 \label{alg:union}
 \begin{algorithmic}[1]
	\REQUIRE $P,Q$\ :SeqBDD
   \ENSURE $P\cup Q$\ :SeqBDD
   \IF {$P= 0\mathchar`- terminal\ node$}
   		\STATE return $Q$
   \ELSIF {$Q= 0\mathchar`- terminal\ node$}
   		\STATE return $P$
   \ELSIF {$P=Q$}
   		\STATE return $P$
   \ELSIF {$(w\leftarrow hash\_table(\langle P,Q\rangle)\ exists)$}
   		\STATE return $w$
   \ENDIF
   \STATE \ \COMMENT {Compare top node's label using alphabetical order}
   \IF {$P.top=Q.top$}
   		\STATE $w\leftarrow Get\_node(P.top,\ P.0\cup Q.0,\ P.1\cup Q.1)$
   \ELSIF {$P.top < Q.top$}
   		\STATE $w\leftarrow Get\_node(P.top,\ P.0\cup Q,\ P.1\cup 0\mathchar`- terminal\ node)$
   \ELSIF {{$P.top > Q.top$}}
   		\STATE $w\leftarrow Get\_node(Q.top,\ P\cup Q.0,\ 0\mathchar`- terminal\ node \cup Q.1)$
   \ENDIF
   \STATE $hash\_table(\langle P,Q\rangle) \leftarrow w$
   \RETURN $w$
 \end{algorithmic}
\end{algorithm}

\section{Twitter pattern extraction example}
\label{sec:apdx_twitter_results}

The following subsections provides four examples for the Twitter evaluation,
good/bad and English/Japanese. 

A slot of a template is indicated by a POS sequence in parenthesis
[\ ] and the number in parenthesis [\ ] at the end of the line shows the usage frequency of the template.

\scriptsize

\subsection{Good Example : English}
Input:@weatherchannel (total 3227 tweets)

\begin{breakbox}
\noindent
-\# Severe thunderstorm watch for portions of western ND and northwestern SD is in effect until 3 a.m. CDT. \# NDwx \# SDwx https://t.co/qOoPdBwWHN\\
-Florida's transition to the wet season is kicking into gear this week: https://t.co/zXCGhxHkoR https://t.co/qodQw0WFqt\\
-In honor of \# MelanomaMonday \&amp; @OutruntheSunInc, stay safe w/these facts about \# melanoma. https://t.co/lNnZzk70C9 https://t.co/BfZW6YuciY\\
-Line of \# severe t-storms w/ high winds, perhaps brief \# tornado, pushing into \# Houston metro. https://t.co/EU10HpTSOK https://t.co/Vm1bj5X5Xk\\
-New \# severe t-storm watch until 11a CT includes \# Houston. Damaging winds main threat. https://t.co/rUFHq68e8h\\
-Rinse...repeat. Yet another setup for local flash \# flooding in \# Texas this week. https://t.co/iqPzZhX386 \# txwx https://t.co/8eYPftAZ5e\\
-Science is about to make cooking a whole lot easier. From chopping to cleaning up, this robot has dinner covered. https://t.co/S0eLUABpa4\\
-Severe T-Storm Watch for portions of Southwest Texas until 10 p.m. CDT. \# TXwx https://t.co/Zizty3vfRH\\
-Tornado Warning for Aitkin, Itasca and St. Louis Counties in MN until 7:30 PM CDT https://t.co/TzipGRIIa5\\
-Tornado Warning for Baltimore and Howard Counties in MD until 2:00 PM EDT https://t.co/GkZEDFI9ga\\
-Tornado Warning for Berkeley and Morgan Counties in WV until 5:00 PM EDT https://t.co/ufrUraPlIP\\
-Tornado Warning for Cheyenne County in CO until 2:45 PM MDT https://t.co/cW9FalG2TX\\
-Tornado Warning for Clay, Jasper and Richland Counties in IL until 8:30 PM CDT https://t.co/f4UT9Vemul\\
-Tornado Warning for Fannin and Lamar Counties in TX until 6:45 PM CDT https://t.co/B6MEBnVNC6\\
-Tornado Warning for Harrison and Marion Counties in TX until 1:45 AM CDT https://t.co/fFHwLqXQ8n\\
-Tornado Warning for Ochiltree and Roberts Counties in TX until 8:45 PM CDT https://t.co/86k9zvmjkY\\
-Tornado Warning for Tillman County in OK until 2:00 AM CDT https://t.co/aC8vnA4PUk\\
-We are tracking more \# snow in the West \&amp; a midweek \# severe threat for the \# Plains...get latest forecast NOW on @AMHQ https://t.co/qUB7LQivQE\\
\end{breakbox}

\vspace{10mm}
Output:@weatherchannel (total 98 patterns)
\begin{breakbox}
\noindent
1:	Tornado Warning for [NNP] County in [NNP] until [CD] PM CDT ( URL )	[440]\\
2:	Tornado Warning for [NNP] , [NNP] , [NNP] and [NNP] Counties in [NNP] until [CD] PM CDT ( URL )	[36]\\
3:	Tornado Warning for [NNP] County County in [NNP] until [CD] PM CDT ( URL )	[33]\\
4:	[VBN]	[9]\\
5:	Tornado Warning for [NNP] County and [NNP] Counties in [NNP] until [CD] PM CDT ( URL )	[9]\\
6:	[IN] [DT] [NN]	[9]\\
7:	[CD] [NNS]	[8]\\
8:	Severe weather threat	[7]\\
9:	LIVE NOW	[7]\\
10:	Tornado of \# 	[6]\\
11:	Tornado [NN] [VBZ]	[6]\\
\end{breakbox}

\subsection{Bad Example (Incorrect Template Obtained) : English}
Input:@virtualemp
\begin{breakbox}
\noindent
-3 Computing Technologies That Will Change the Face of Business World – http://t.co/f2KeWQbHRc\\
-5 project management hacks to make your project see the light of day – http://t.co/TgpWAFVHon\\
-5 Secrets to Increase \# Employee Engagement With Technology https://t.co/Jq8Y0q6iWG \# retention  \# work \# business https://t.co/gN110xSUk4\\
-6 Consumer Psychology Hacks You Need to Know NOW! - Neuromarketing https://t.co/uFXBkUoSCe \# socialmedia \# marketing \# smallbiz\\
-A freelancer can be a handy resource when the work you have is short-term and low-budget  http://t.co/g9aYxXLh6U\\
-Apart from cost effectiveness, there are certain pertinent advantages of \# outsourcing.\\
-Facebook Acquires Online Shopping Search Engine TheFind – http://t.co/56GUwUyVh7\\
-Google’s SDKs Enable iOS, Android Mobile Apps to Work Offline – http://t.co/F2BvkuvNfs\\
-Hire Full Time Remote C++ Programmers To Work Dedicatedly For You:  http://t.co/WGFQIeSzOB\\
-Hiring remote Photoshop experts is the best way to get customized solutions since these are your very own resources. http://t.co/yyKsIcObJ4\\
-How To Manage Your Remote Development Team https://t.co/omAllH2DWd \# business \# development \# management https://t.co/iqxovtPBKv\\
-How to measure the success of your digital marketing campaign. Find out at http://t.co/aQAWHndmEL\\
-How to Merge your Business Processes with Online Marketing – http://t.co/ifIqwGcGES\\
-How to navigate the dynamic digital marketing world https://t.co/j4FuDd0B12 \# marketing \# socialmedia \# smallbiz\\
-http://t.co/Rqf6R7vZGH Benefits of hiring dedicated Blackberry application developers with http://t.co/SiomE2CdQk\\
-Infosys shares surge 9\% on founder Narayana Murthy's return  http://t.co/KdJrzW8KdB\\
-Let’s check out the facts why a remote dedicated payroll expert is a better resource than a freelancer.  http://t.co/f1LMM5vBIa\\
-Microsoft and Amazon leading cloud computing performance: Edmund Shing- https://t.co/vUWrVR7OjV\\
-Must add tools for your \# startup \# enterprenuer  https://t.co/FauI57Abg7\\
-Outsourcing Helps You Recruit Professional Lawyers https://t.co/T7n8jW2L63 \# Outsource \# Lawyers \# smallbusiness\\
-Outsourcing Human Resources: The Best Business Decision You Will Ever Make. Read on – http://t.co/NCHkgYjpxQ\\
-Outsourcing Human Resources: The Best Employment Decision You Will Ever Make. Find out why this is the case. Click http://t.co/Ugv8izBZeQ\\
-Philippines’ outsourcing dream crashes – unable to meet talent demand http://t.co/YEIWXXPwR7 \# outsourcingtrends\\
-Precise information about Dot Net team model of http://t.co/9z6sNop6jp, have a look! http://t.co/ChW5TtIaQ2\\
-Procter \&amp; Gamble is doing a major shift in marketing as over 100 brands will be taken off the shelves: http://t.co/W9gsSRzcsx\\
-The benefits of hiring a remotely working dedicated interior designer:  http://t.co/0CeJKwfDSL\\
-Twitter Inks Deal with Ad Giant WPP to Expand Data-Driven Marketing:  http://t.co/NcB4ZK3Whq\\
-VE Opens Its Third Office – in Birmingham, UK http://t.co/9C8VtWVHbI \# whatsnewatve\\
-Virtual Employee Pvt Ltd, an innovative outsourcing platform has been awarded ISO 9001:2008 Certification. http://t.co/sC9XShsLkU\\
-Why Is a Dedicated Remote Resource the Best Way to Outsource Your Financial Services?: http://t.co/4XMDIqVROC\\
-Why is India the best outsourcing destination for website development? http://t.co/Lsgit0QMP1\\
\end{breakbox}

\vspace{10mm}
Output:@virtualemp
\begin{breakbox}
\noindent
1: [VBG] to [11]\\
2: [NNP] ,  [11]\\
3: The [JJ] [NN] [NN]   [10]\\
4: [NNP] [VBZ] [JJ]  [9]\\
5: Did   [9]\\
6: [NNS] new   [8]\\
7: How [PRP] [MD] [VB]  [8]\\
8: [NNP] [VBZ] [NNP] [8]\\
9: [CD] [JJ] [NNS] to [VB] [7]\\
10:   [VBP] you [VBG]   [7]\\

\end{breakbox}

\normalsize